\documentclass[final,5p,times,twocolumn,sort&compress]{elsarticle}

\usepackage{subfig}
\usepackage[labelfont=bf,labelsep=period]{caption}
\usepackage{microtype}
\usepackage{graphicx}
\usepackage{booktabs} 
\usepackage{hyperref}
\hypersetup{colorlinks=true, 
            urlcolor=urlcolor, 
            linkcolor=linkcolor, 
            anchorcolor=anchorcolor, 
            citecolor=citecolor}

\usepackage{amsmath}
\usepackage{amssymb}
\usepackage{mathtools}
\usepackage{amsthm}
\usepackage{bm}

\usepackage{newtxmath}
\usepackage{makecell}

\usepackage[capitalize,noabbrev]{cleveref}
\usepackage[dvipsnames]{xcolor}
\usepackage{graphicx}

\usepackage{epsfig}
\usepackage{makecell}

\usepackage{setspace}
\usepackage{appendix}
\usepackage{titletoc}

\usepackage{multirow}
\usepackage{caption}
\usepackage{hyperref}
\usepackage{lipsum}

\usepackage[utf8]{inputenc} 
\usepackage[T1]{fontenc}    
\usepackage{url}            
\usepackage{booktabs}       
\usepackage{amsfonts}       
\usepackage{nicefrac}       
\usepackage{microtype}      
\usepackage{colortbl}
\usepackage{array}
\usepackage{pifont}

\definecolor{backward}{HTML}{E7E7E7}
\definecolor{aliceblue}{rgb}{0.94, 0.97, 1.0}
\definecolor{grey}{rgb}{0.5,0.5,0.5}
\definecolor{blues}{rgb}{0.019,0.388,0.756}
\definecolor{reddot}{rgb}{0.87,0.17,0.17}
\definecolor{cont_green}{rgb}{0.6,0.8,0.6}
\definecolor{cont_red}{rgb}{0.8,0.34,0.34}
\definecolor{citecolor}{HTML}{0071BC}
\definecolor{linkcolor}{HTML}{ED1C24}
\definecolor{urlcolor}{HTML}{004D99}
\definecolor{anchorcolor}{HTML}{555555}

\newcommand{\eg}[0]{{e.g.}}
\newcommand{\ie}[0]{{i.e.}}

\newcommand{\R}{\mathbb{R}}

\def\sG{{\mathbb{G}}}

\def\sI{{\mathbb{I}}}

\makeatletter
\def\ps@pprintTitle{%
 \let\@oddhead\@empty
 \let\@evenhead\@empty
 \def\@oddfoot{}%
 \let\@evenfoot\@oddfoot}
\makeatother

\makeatletter
\renewcommand{\paragraph}{%
  \@startsection{paragraph}{4}%
  {\z@}{0ex \@plus 0ex \@minus 0ex}{-1em}%
  {\hskip\parindent\normalfont\normalsize\itshape}%
}
\makeatother

\journal{Engineering}

\begin{document}
\thispagestyle{empty}
\begin{frontmatter}

\title{CktGen: Automated Analog Circuit Design with Generative Artificial Intelligence}

\author[zju_cs]{Yuxuan Hou}
\ead{yuxuan.hou.x@gmail.com}
\author[zju_cs]{Hehe Fan\corref{cor}}
\ead{hehe.fan.cs@gmail.com}
\author[uts]{Jianrong Zhang}
\author[zju_cs]{Yue Zhang}
\author[zju_ee]{Hua Chen\corref{cor}}
\ead{chenhua@zju.edu.cn}
\author[zju_ee]{Min Zhou}
\author[zju_ee]{Faxin Yu}
\author[nus]{Roger Zimmermann}
\author[zju_cs]{Yi Yang}

\cortext[cor]{Corresponding authors}

\affiliation[zju_cs]{
    organization={College of Computer Science and Technology},
    addressline={Zhejiang University}, 
    postcode={Hang Zhou 310027}, 
    country={China}
}

\affiliation[uts]{
    organization={Australian Artificial Intelligence Institute},
    addressline={University of Technology Sydney}, 
    postcode={Ultimo NSW 2007}, 
    country={Australia}
}

\affiliation[zju_ee]{
    organization={School of Aeronautics and Astronautics},
    addressline={Zhejiang University}, 
    postcode={Hang Zhou 310027}, 
    country={China}
}

\affiliation[nus]{
    organization={School of Computing},
    addressline={National University of Singapore}, 
    postcode={Singapore 117417}, 
    country={Singapore}
}

\begin{abstract}
The automatic synthesis of analog circuits presents significant challenges. Most existing approaches formulate the problem as a single-objective optimization task, overlooking the fact that design specifications for a given circuit type can vary widely across applications. To address this limitation, we introduce specification-conditioned analog circuit generation, a task that directly generates analog circuits based on stated specifications. The motivation is to find an effective method that leverages existing well-designed circuits to improve automation in analog circuit design. Specifically, we propose CktGen, a simple yet effective variational autoencoder model that maps discretized specifications and circuits into a joint latent space and reconstructs the circuit from that latent vector. Notably, as a single specification may correspond to multiple valid circuits, naively fusing the specification information into the generative model does not capture these one-to-many relationships. To address this, we first decouple the encoding process of circuits and specifications and align their mapped latent space. Then, we employ contrastive training with a filter mask to maximize differences between encoded circuits and specifications. Furthermore, classifier guidance along with latent feature alignment promotes the clustering of circuits sharing the same specification, thus avoiding model collapse into trivial one-to-one mappings. By canonicalizing the latent space with respect to the specifications, we can further optimize and search for an optimal circuit that meets the valid target specification. We conduct comprehensive experiments on the open circuit benchmark and introduce several metrics to evaluate cross-model consistency in the specification-conditioned circuit generation task. The experimental results demonstrate that CktGen achieves substantial improvements over existing state-of-the-art methods.

\end{abstract}


\end{frontmatter}

\section{Introduction}

\par Analog circuits are essential for processing continuous signals, but their design remains labor-intensive and highly dependent on expert intuition. Automated design aims to bridge this gap by directly mapping target specifications to viable circuit implementations. Despite the apparent correspondence between specifications and realizable circuits, modeling this complex, high-dimensional mapping remains a major challenge. As circuit complexity grows, there is an urgent need for automated tools to accelerate the analog design process.

\par Analog circuit synthesis involves two principal tasks: topology selection and device sizing~\cite{design_flow_survey}. The objective of analog circuit synthesis is to ensure that the circuit meets the target specifications and achieves better trade-offs among them. Most previous studies treat analog circuit synthesis as an optimization problem~\cite{survey1, survey2, survey3}. One optimization goal is to maximize the figure of merit~(FoM), a metric that quantifies the overall trade-offs between circuit specifications (such as gain, bandwidth, and phase margin) within given constraints~\cite{syn_heu_1, sz_rl_1, cktgnn}. Another optimization objective is to ensure that the performance of the synthesized circuits meets the target specifications~\cite{topo_rule_2, syn_heu_2}. Existing approaches can be broadly categorized as knowledge-based~\cite{topo_rule_1, sz_know_1}, learning-based~\cite{sz_rl_1, topo_rl_1}, and simulation-based methods~\cite{topo_rl_2, topo_rl_3}.

\begin{figure*}[t!]
\centering
\centerline{\includegraphics[width=0.8\textwidth]{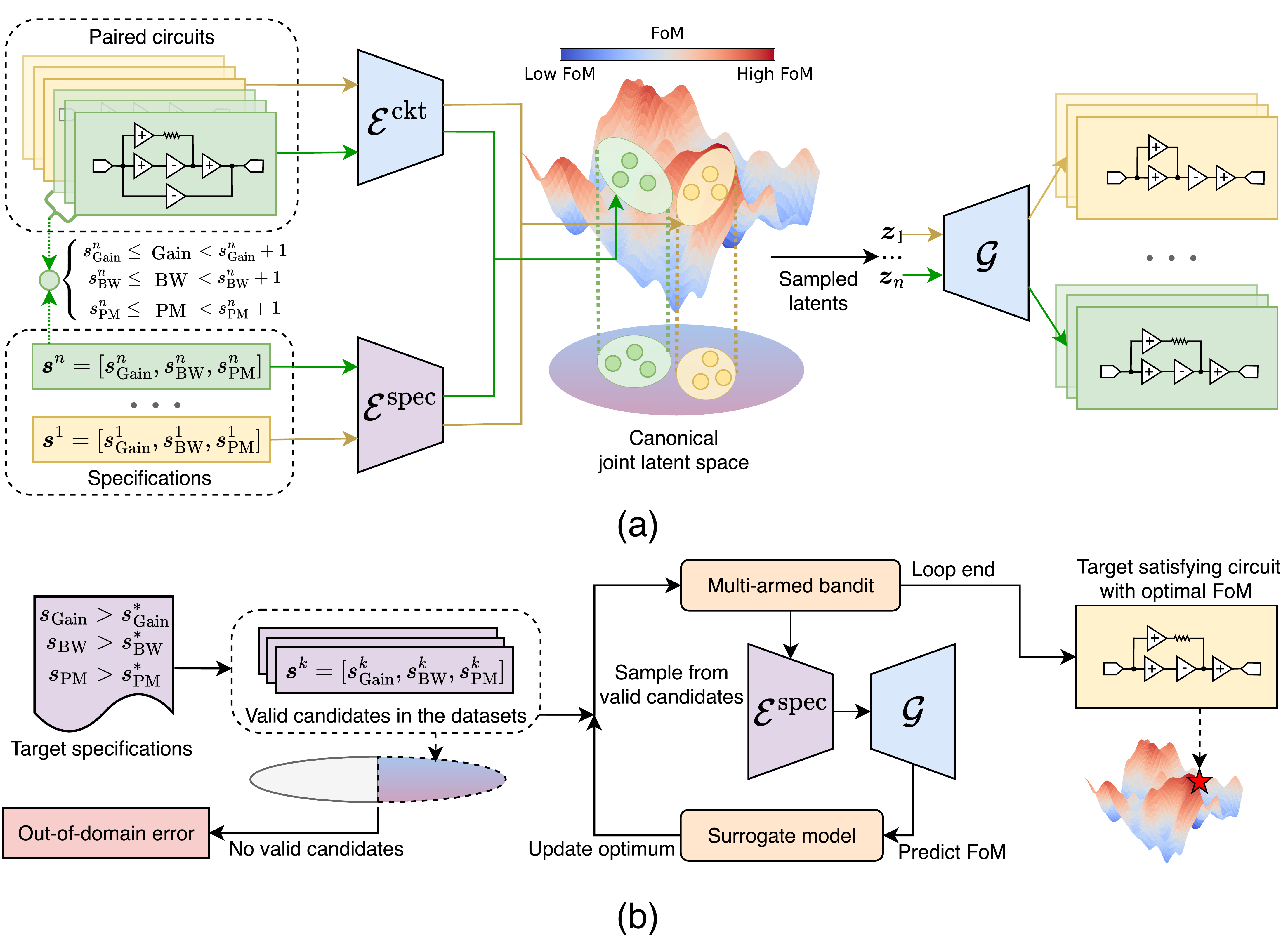}}
\caption{
    Overview of the CktGen framework for specification-conditioned analog circuit generation and optimization.
    (a) Joint representation learning. Performance specifications (gain $s_{\text{Gain}}$, bandwidth $s_{\text{BW}}$, and phase margin $s_{\text{PM}}$) are discretized into interval-based classes, grouping circuits by their joint specification class $\bm{s}$. Dedicated encoders map circuits and specifications into a canonical joint latent space. Contrastive training and classifier guidance enforce discriminative boundaries between classes while preserving those within-class cohesion, enabling one-to-many mapping from specifications to diverse circuits. The circuit generator $\mathcal{G}$ synthesizes circuits from specification latent vectors.
    (b) Test-time optimization. The specific design requirements are modeled as inequality constraints defined by target specification thresholds $\bm{s}^*$ for gain, bandwidth, and PM (i.e., $s_{\text{Gain}} > s^*_{\text{Gain}}$, $s_{\text{BW}} > s^*_{\text{BW}}$, and $s_{\text{PM}} > s^*_{\text{PM}}$), defining a feasible region in the learned latent space. The framework identifies valid joint specification classes that overlap with or approximate this feasible region, reporting an out-of-domain error if none exist. For valid cases, a Bayesian multi-armed bandit~(MAB) algorithm performs test-time optimization without model retraining, iteratively searching the latent space, sampling candidates, generating circuits via $\mathcal{G}$, and evaluating their figure of merit (FoM) via a surrogate model. The algorithm adaptively refines its search strategy based on observed FoM values, converging to an optimal circuit that meets the target specification.
        $\mathcal{E}^{\text{ckt}}$:  a transformer-based variational autoencoder for circuits; 
        $\mathcal{E}^{\text{spec}}$: a multilayer perceptron encoder for specifications; 
        $\bm{s}^1, \ldots, \bm{s}^n$: all the joint specification classes in the training dataset, $n$: the is number of the joint specification classes in the dataset range; 
        $\bm{z}_1,\ldots,\bm{z}_n$: latent vectors sampled for each joint specification class; 
        $s^{*}_{\text{Gain}}$, $s^{*}_{\text{BW}}$, and $s^{*}_{\text{PM}}$: target specification thresholds for gain, bandwidth and phase margin, respectively; 
        $\bm{s}^k$: the $k$-th valid specification candidate in the dataset range that meets the threshold ($k=1, 2, \ldots, k_{\text{total}}$, with $k_{\text{total}}$ being the total number of available candidates); 
        $s^{k}_{\text{Gain}}$, $s^{k}_{\text{BW}}$, and $s^{k}_{\text{PM}}$: the gain, bandwidth, and phase margin of the $k$th valid specification class candidate, respectively.
}
\label{fig:intro}
\end{figure*}

\par Knowledge-based approaches use domain rules~\cite{topo_rule_2, topo_rule_1} or analytical models~\cite{sz_know_1, sz_know_2}, perform well for simple circuits~\cite{topo_rule_3, topo_rule_4, sz_know_3} but do not scale to more complex designs. To improve scalability, heuristic algorithms have been applied to automate both topology generation~\cite{topo_heu_2, topo_heu_3, topo_heu_5, topo_heu_7} and device sizing~\cite{sz_heu_2, sz_heu_3}. However, such methods are highly dependent on initialization and the performance is often unpredictable. More recently, machine learning has enabled new progress. Graph-based generative models~\cite{topo_bo} and reinforcement learning~\cite{topo_rl_1, topo_rl_2} allow more efficient circuit exploration, while Bayesian optimization~\cite{sz_bo_1, sz_bo_2} and surrogate models~\cite{cktgnn, sz_nn_1, sz_nn_2, surrogate} help reduce reliance on expensive simulations. Deep learning also enhances circuit analysis, such as by enabling better feature extraction~\cite{review_conv} or fault detection~\cite{review_vq}. While simulation-based methods using tools such as simulation program with integrated circuit emphasis (SPICE)~\cite{spice} remain the standard for accuracy, they are computationally costly and are often combined with faster learning or heuristic models~\cite{sz_rl_3}.

\par Despite these advances, most existing approaches treat topology selection and device sizing as separate steps. Some recent methods aim to address both tasks jointly, using genetic programming~\cite{syn_heu_1}, integrated pipelines~\cite{syn_heu_2, wicked}, or neural network models~\cite{cktgnn, dvae, syn_lrn_1}. For certain specific circuit types, classical sizing techniques are combined with automated topology searching~\cite{syn_lrn_2, gm/id, magical}. Yet, most solutions cannot easily adapt to changing requirements~\cite{sz_rl_1, topo_rl_3, sz_bo_3, sz_rl_2, syn_lrn_3}. A generalizable approach that can automatically generate circuits from changing specifications remains a major open challenge.

\par A key limitation of current approaches is their reliance on a small number of manually predefined specification settings. Most methods either explicitly optimize for these fixed target cases or only focus on maximizing the FoM~\cite{cktgnn, topo_rl_2, topo_rl_3, topo_bo}. In practice, design requirements often shift, requiring new specifications even for the same circuit type. Existing methods can only address a limited set of predefined target cases and require additional training or complete re-optimization when specifications change, severely restricting their generalization and flexibility. To overcome these limitations, we reformulate analog circuit synthesis as a conditional generation problem, enabling the model to capture one-to-many relationships between specifications and circuit implementations from existing design data. This paradigm allows the model to accommodate shifting requirements and support a broader spectrum of design scenarios through test-time optimization without retraining.

\par In this work, we introduce CktGen, a conditional generative framework for automated analog circuit design. As illustrated in Fig.~\ref{fig:intro}(a), to facilitate joint representation learning for specifications and circuits, we discretize performance specifications (gain $s_{\text{Gain}}$, bandwidth $s_{\text{BW}}$, and phase margin (PM) $s_{\text{PM}}$) into intervals, grouping circuits by their joint specification class $\boldsymbol{s}=[s_{\text{Gain}}, s_{\text{BW}}, s_{\text{PM}}]$. To align circuits and specifications, CktGen maps both specifications and paired circuits into a canonical joint latent space, using a transformer-based variational autoencoder~(VAE) \cite{vae} for circuits~($\mathcal{E}^{\text{ckt}}$) and a multilayer perceptron encoder for specifications $(\mathcal{E}^{\text{spec}}$). We structure this space using contrastive learning~\cite{infonce, simclr_v2, moco_v2, zhang2022region} and classifier guidance with feature alignment, which enforces discriminative boundaries between classes while preserving within-class cohesion. This enables CktGen to learn robust one-to-many mappings from specifications to candidate circuits. Given a valid discretized specification, the circuit generator $\mathcal{G}$ produces diverse corresponding circuits. As shown in Fig.~\ref{fig:intro}(b), at test time, the specific design requirements are modeled as inequality constraints defined by target specification thresholds $\bm{s}^*$ for gain, bandwidth, and PM (i.e., $s_{\text{Gain}} > s^*_{\text{Gain}}$, $s_{\text{BW}} > s^*_{\text{BW}}$, and $s_{\text{PM}} > s^*_{\text{PM}}$), defining a feasible region in the latent space. The framework identifies valid joint specification classes that overlap with this feasible region, then performs test-time optimization without model retraining, employing a multi-armed bandit~(MAB) algorithm to efficiently search for high-performance circuits in the learned latent space. This framework generates and optimizes designs for diverse specified requirements without retraining or restriction to preset cases.

\par To evaluate the generated circuits, we train a graph isomorphism network (GIN)-based~\cite{gin} surrogate model with cross-modal alignment losses to verify whether each candidate meets its target specifications and to predict its FoM value. We also introduce new metrics to systematiclly measure how well the generated circuits match their specifications, addressing an important gap in previous work. Our approach achieves substantial improvements in conditional circuit generation, automated design, reconstruction, and unconditional generation tasks.

\par In summary, our contributions are as follows:
\begin{itemize}
\item We present a specification-conditioned generation framework for analog circuit synthesis that overcomes the flexibility limitation of traditional optimization-based approaches, enabling adaptation to changing design requirements without retraining.
\item We employ contrastive training and classifier guidance with feature alignment to align circuits and specifications in a canonical joint latent space. Our CktGen learns a clustered representation that naturally supports one-to-many mappings from specifications to diverse circuits. 
\item We introduce test-time optimization using an MAB algorithm, which efficiently searches for high-performance circuits without model retraining.
\item We introduce a few metrics to quantitatively assess the consistency between generated circuits and their specifications, filling a gap in previous evaluation methods. Our model significantly outperforms previous methods across all tasks.
\end{itemize}

\section{Methods}
\subsection{Task statement}
\label{sec:task}
\par The specification-conditioned analog circuit generation task aims to generate analog circuits based on given specifications. More specifically, given $\bm{s}=[s_{\text{Gain}}, s_{\text{BW}}, s_{\text{PM}}]$, the analog circuit is expected to be generated from $\bm{s}$. To achieve this, we preprocess the specifications from datasets, including Ckt-Bench-101 and Ckt-Bench-301~\cite{cktgnn}. We first discard the data with invalid specifications~(\ie, PM $<$ 0). Next, we simplify the specifications by truncating the fractional parts and categorizing them. Finally, each specification element is processed into a one-hot representation. The ranges and discretized class counts of the processed data are detailed in Table~\ref{spec_data}, and the data distribution is shown in \ref{appd:sec:a}.

\par The circuit is represented as a directed acyclic graph~(DAG) $G(V, E)$, where $V$ denotes the nodes and $E$ represents the directed edges. Note that the specifications are graph-level features. The node feature matrix is denoted as $\bm{X} = [\bm{x}_1, \cdots, \bm{x}_N]^{\top}$, where $N$ is the number of nodes in the DAG. Specifically, the feature vector for the $i$th node is defined as $\bm{x}_i = [x^{\text{type}}_i, x^{\text{pos}}_i, \bm{x}^{\text{size}}_i] (i=1, 2, \ldots, N)$. Here, $x^{\text{type}}_i$ denotes the type feature of the $i$th node, selected from 26 subgraph categories; details are provided in \ref{appd:sec:b}. $x^{\text{pos}}_i$ represents the node position feature, which indicates whether the node is in the main path of the circuit~(\ie, a feedforward path where the direction flows from the input node to the output node). $\bm{x}^{\text{size}}_i$ is a vector representing the device parameters for the $i$th node. The edges $E$ are represented as the adjacency matrix $\bm{A} \in \R^{N\times N}$ and the flattened edge list $\bm{x}^{\text{edge}} \in \R^{N\times (N-1)/2}$, where $\R$ is the set of real numbers.

\begin{figure*}[t!]
    \centering
    \centerline{\includegraphics[width=\textwidth]{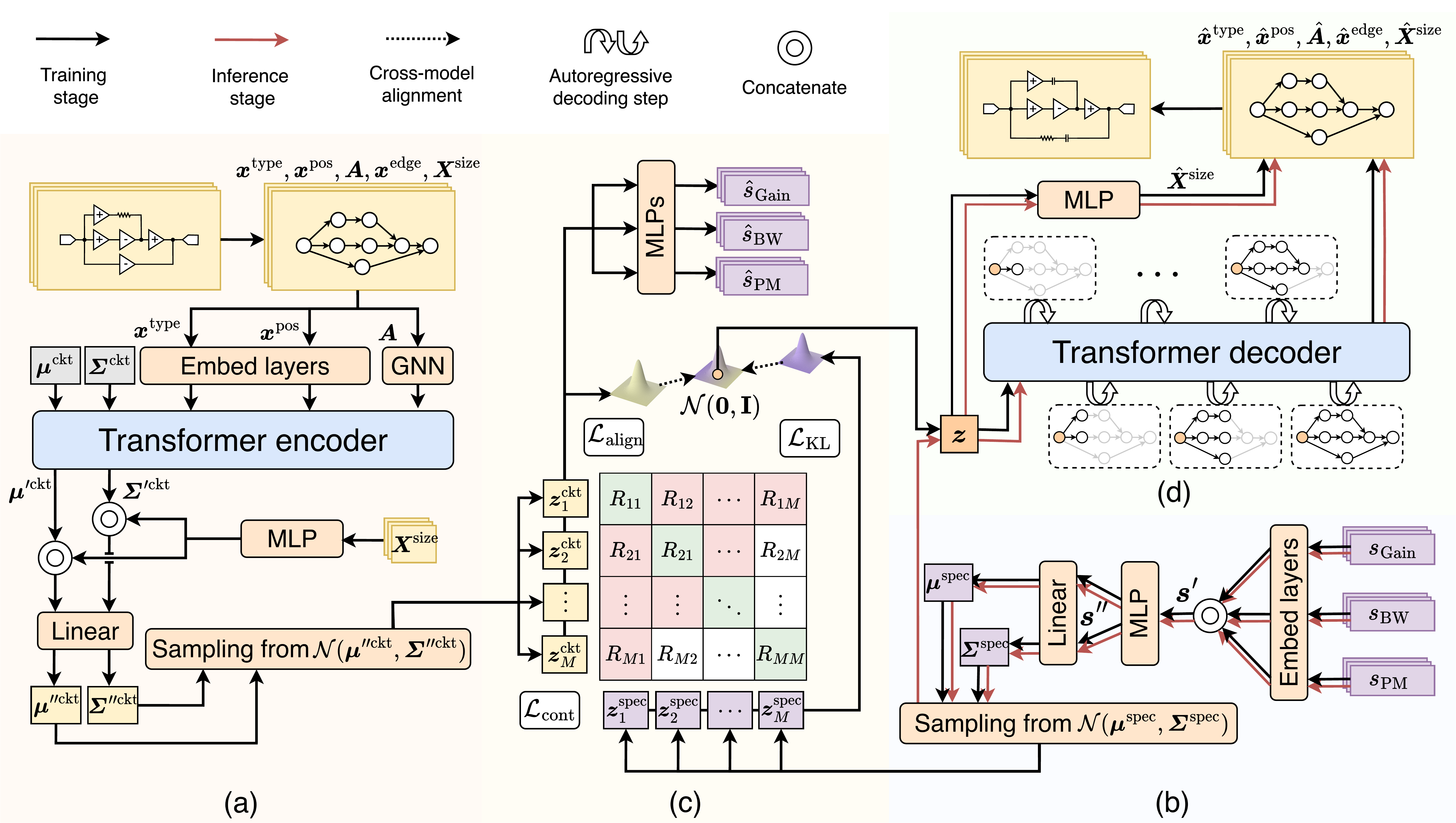}}
    \caption{Overview of the CktGen architecture.
    The model consists of four modules: 
    (a) a circuit encoder $\mathcal{E}^{\text{ckt}}$, (b) a specification encoder $\mathcal{E}^{\text{spec}}$, (c) latent space alignment, and (d) a circuit generator $\mathcal{G}$. The circuit encoder and specification encoder map circuits and specifications into a latent space. To cluster circuits with the same specification and distinguish different ones, we employ contrastive training and classifier guidance to train a joint latent space. In the circuit generator, a generative pre-trained transformer~(GPT)-like model generates the features of the circuit autoregressively. The definitions of the variables in the figure can be found in Sections~\ref{sec:encoder} and Sections~\ref{sec:align}.
    }
    \label{fig:framework}
\end{figure*}

\begin{table}[t!]
    \centering
    \begin{scriptsize}
        \caption{Value ranges and discretized class counts in the processed datasets.}
        \begin{tabular}{lcccccc}
        \hline
        & \multicolumn{3}{c}{Ckt-Bench-101} & \multicolumn{3}{c}{Ckt-Bench-301} \\
        & \multicolumn{3}{c}{dataset attribute} & \multicolumn{3}{c}{dataset attribute} \\ \cline{2-7}

        \multirow{-2.9}{*}{Item} & $s_{\text{Gain}}$ & $s_{\text{BW}}$ & $s_{\text{PM}}$ & $s_{\text{Gain}}$ & $s_{\text{BW}}$ & $s_{\text{PM}}$\\ \hline
        Valid range & $[0, 4)$ & $[0, 32)$ & $[0, 6)$ & $[0, 4)$ & $[0, 19)$ & $[0, 5)$ \\
        Discrete class count & $4$ & $32$ & $6$ & $4$ & $19$ & $5$  \\ \hline 
        \end{tabular}
        \label{spec_data}
    \end{scriptsize}
\end{table}

\subsection{Architecture of CktGen}
\label{sec:archi}
\par CktGen is a conditional VAE model with three main components: a circuit encoder, specification encoder, and circuit generator. The overall architecture is illustrated in Fig.~\ref{fig:framework}. The circuit encoder and specification encoder map circuits and specifications into a latent space, and the circuit generator reconstructs the circuits from the latent vectors. Furthermore, contrastive training and classifier guidance with feature alignment are used to construct a high-quality joint latent space of the circuits and specifications.

\subsubsection{Circuit encoder and specification encoder}
\label{sec:encoder}
\par We introduce two encoders to map the circuits and their specifications into latent vectors. Our goal is to align the latent representations from these two modalities. Two different architectures are considered for these two modalities. For the circuits (as shown in Fig.~\ref{fig:framework}(a)), the inputs include the node type feature vector $\bm{x}^{\text{type}}=[x_{1}^{\text{type}},\ldots,x_{N}^{\text{type}}]$, the node position feature vector $\bm{x}^{\text{pos}}=[x_{1}^{\text{pos}},\ldots,x_{N}^{\text{pos}}]$, the device parameter feature matrix $\bm{X}^{\text{size}}=[\bm{x}_{1}^{\text{size}},\ldots,\bm{x}_{N}^{\text{size}}]$, an adjacency matrix $\bm{A}$, and two learnable distribution tokens $\bm{\mu}^{\text{ckt}}$ and $\bm{\mathnormal{\Sigma}}^{\text{ckt}}$. The node feature vectors $\bm{x}^{\text{type}}$ and $\bm{x}^{\text{pos}}$ are projected into embedding $\bm{x}'^{\text{type}} \in \mathbb{R}^{N \times d}$ and $\bm{x}'^{\text{pos}} \in \mathbb{R}^{N \times d}$, respectively, where $d$ is the dimension of the node feature embedding. We feed $\bm{A}$ into a graph neural network (GNN)~\cite{gnn} to obtain connectivity embedding $\bm{A}' \in \mathbb{R}^{N \times d}$. Then, we use a transformer-based VAE $\mathcal{E}^{\text{ckt}}$ to obtain $\bm{\mu}'^{\text{ckt}}$ and $\bm{\mathnormal{\Sigma}}'^{\text{ckt}}$, which can be computed as follows:
\begin{small}
\begin{equation}
    \bm{\mu}'^{\text{ckt}}, \bm{\mathnormal{\Sigma}}'^{\text{ckt}} = \mathcal{E}^{\text{ckt}}(\bm{\mu}^{\text{ckt}}, \bm{\mathnormal{\Sigma}}^{\text{ckt}}, \bm{x}'^{\text{type}}, \bm{x}'^{\text{pos}}, \bm{A}')
\end{equation}
\end{small}%
\noindent where $\bm{\mu}'^{\text{ckt}}$ and $\bm{\mathnormal{\Sigma}}'^{\text{ckt}}$ represent the learned distribution parameters from the circuit encoder. Furthermore, $\bm{X}^{\text{size}}$ is projected using an MLP and then concatenated with $\bm{\mu}'^{\text{ckt}}$ and $\bm{\mathnormal{\Sigma}}'^{\text{ckt}}$ separately. Finally, two fully connected layers are used to obtain the final distribution parameters $\bm{\mu}''^{\text{ckt}}$ and $\bm{\mathnormal{\Sigma}}''^{\text{ckt}}$, which incorporate the device parameters.

\par For the specifications (as shown in Fig.~\ref{fig:framework}(b)), we take $\bm{s}$ as input. Each specification (\ie, $s_{\text{Gain}}$, $s_{\text{BW}}$, and $s_{\text{PM}}$) is first projected into the $\mathbb{R}^{d}$ space, obtaining embedded specifications $\bm{s}'_{\text{Gain}}$, $\bm{s}'_{\text{BW}}$, and $\bm{s}'_{\text{PM}}$. We then concatenate them to form a unified vector $\bm{s}' \in \mathbb{R}^{3d}$ and employ an MLP encoder $\mathcal{E}^{\text{spec}}$ to obtain the joint specification representation $\bm{s}''\in \mathbb{R}^{d}$, which can be written as follows:
\begin{small}
\begin{equation}
    \bm{s}'' = \mathcal{E}^{\text{spec}} \left(\mathrm{concat} \left(\bm{s}'_{\text{Gain}}, \bm{s}'_{\text{BW}}, \bm{s}'_{\text{PM}} \right) \right)
\end{equation}
\end{small}%
Finally, two fully connected layers are used to extract the specification distribution parameters $\bm{\mu}^{\text{spec}}$ and $\bm{\mathnormal{\Sigma}}^{\text{spec}}$, respectively.

\par By applying the reparameterization trick~\cite{vae}, we sample the circuit latent vector $\bm{z}^{\text{ckt}}  \in \mathbb{R}^{d'}$ and the specification latent vector $\bm{z}^{\text{spec}} \in \mathbb{R}^{d'}$ from their respective learned distribution parameters (i.e., $\{\bm{\mu}''^{\text{ckt}}, \bm{\mathnormal{\Sigma}}''^{\text{ckt}}\}$ and $\{\bm{\mu}^{\text{spec}}, \bm{\mathnormal{\Sigma}}^{\text{spec}}\}$), where $d'$ denotes the dimension of the latent vector. 

\paragraph{Optimization goal} 
\par To optimize the two encoders, we employ four Kullback-Leibler (KL) losses inspired by TEMOS~\cite{temos}. More specifically, we minimize the KL divergences between the outputs of the two encoders and the normal distribution $\bm{\beta} = \mathcal{N}(\boldsymbol{0}, \bm{\mathrm{I}})$, where $\mathcal{N}$ indicates the normal family of distributions, $\boldsymbol{0}$ represents the zero mean vector, and $\bm{\mathrm{I}}$ represents the identity covariance matrix. Additionally, we use the KL losses to refine the alignment between the output features of the two modalities. The overall KL loss~($\mathcal{L}_{\mathrm{KL}}$) is as follows:
\begin{small}
\begin{equation}
\begin{aligned}   
        \mathcal{L}_{\mathrm{KL}} = 
        &\mathrm{KL}\left(\bm{\alpha}^{\text{ckt}}, \bm{\beta}\right)
        + \mathrm{KL}\left(\bm{\alpha}^{\text{spec}}, \bm{\beta}\right) \\
        + &\mathrm{KL}\left(\bm{\alpha}^{\text{ckt}}, \bm{\alpha}^{\text{spec}}\right) 
        + \mathrm{KL}\left(\bm{\alpha}^{\text{spec}}, \bm{\alpha}^{\text{ckt}}\right)
\end{aligned}
\end{equation}
\end{small}%
\noindent where $\bm{\alpha}^{\text{ckt}} = \mathcal{N}(\bm{\mu}''^{\text{ckt}}, \bm{\mathnormal{\Sigma}}''^{\text{ckt}})$ and $\bm{\alpha}^{\text{spec}} = \mathcal{N}(\bm{\mu}^{\text{spec}}, \bm{\mathnormal{\Sigma}}^{\text{spec}})$ denote the distribution of the circuit and specification latent vectors, respectively.

\subsubsection{Specification-circuit alignment}
\label{sec:align}
\par A single specification usually corresponds to multiple circuits. We observe that simply using the KL loss to align the latent vectors cannot capture these one-to-many relationships~(Section~\ref{sec:gen}). To address this issue, we leverage contrastive training, classifier guidance, and feature alignment to improve the joint latent space.

\paragraph{Contrastive training}
\par We incorporate contrastive training to emphasize the distinctions between circuit latent vectors corresponding to different specifications~(\ie, negative pairs). Consider a batch containing $M$ positive pairs $(\bm{z}^\text{spec}_1, \bm{z}^\text{ckt}_1), (\bm{z}^\text{spec}_2, \bm{z}^\text{ckt}_2),\ldots, (\bm{z}^\text{spec}_M, \bm{z}^\text{ckt}_M)$, where any pair $(\bm{z}^\text{spec}_{p}, \bm{z}^\text{ckt}_{q})$ for $p \neq q$ ($p=1,2,\ldots, M$; $q=1,2,\ldots, M$) is treated as negative. We calculate a similarity matrix $R_{pq} = \cos(\bm{z}^\text{spec}_{p}, \bm{z}^\text{ckt}_{q})$ to capture the cosine similarities among pairs. Subsequently, we apply the information noise-contrastive estimation~(InfoNCE) loss ($\mathcal{L}_{\text{cont}}$)~\cite{infonce} to optimize the circuit encoder and specification encoder, which can be defined as follows:
\begin{small}
\begin{equation}
    \mathcal{L}_{\text{cont}} = -\frac{1}{2M} 
    \sum_p
    \left(\log \frac{e^{R_{pp} / \tau}}{\sum_q e^{R_{pq} / \tau}} + \log \frac{e^{R_{pp} / \tau}}{\sum_q e^{R_{qp} / \tau}}\right)
\end{equation}
\end{small}%
where $\tau$ is the temperature hyperparameter. As depicted in Fig.~\ref{fig:framework}(c), diagonal pairs (colored \textcolor{cont_green}{green}) represent positive latent vector pairs, while off-diagonal pairs (colored \textcolor{cont_red}{red}) represent negatives. It should be noted that we treat circuits that share the same specifications but differ in topology and device parameters as negatives (colored white) and filter them out during loss computation.

\paragraph{Classifier guidance}
\par We propose a classifier guidance loss ($\mathcal{L}_{\text{guide}}$) to further improve the cross-modal consistency in the latent space. We feed the encoded circuit latent vector $\bm{z}^{\text{ckt}}$ into three MLPs to predict gain, bandwidth, and phase margin, thereby obtaining the predicted specifications $\hat{s}_{\text{Gain}}$, $\hat{s}_{\text{BW}}$ and $\hat{s}_{\text{PM}}$, respectively. Then, $\mathcal{L}_{\text{guide}}$ can be computed with the ground-truth specifications as follows:
\begin{small}
\begin{equation}
\label{eq_cg}
    \mathcal{L}_{\text{guide}} = 
    \mathcal{L}_{\text{CE}} (s_{\text{Gain}}, \hat{s}_{\text{Gain}}) + 
    \mathcal{L}_{\text{CE}} (s_{\text{BW}}, \hat{s}_{\text{BW}})
    + \mathcal{L}_{\text{CE}} (s_{\text{PM}}, \hat{s}_{\text{PM}})
\end{equation}
\end{small}%
\noindent where $\mathcal{L}_{\text{CE}}$ denotes the cross-entropy loss.

\paragraph{Feature alignment}
Following TEMOS~\cite{temos}, we further encourage consistency between $\bm{z}^{\text{ckt}}$ and $\bm{z}^{\text{spec}}$:
\begin{small}
\begin{equation}
    \mathcal{L}_{\text{feat}} = \mathcal{L}_{1}\left(\bm{z}^{\text{ckt}}, \bm{z}^{\text{spec}}\right)
\end{equation}
\end{small}%
\noindent where $\mathcal{L}_{\text{feat}}$ denotes the latent feature alignment loss that minimizes the distance between the circuit latent vector and the specification latent vector, and $\mathcal{L}_{1}$ denotes the smooth $\mathcal{L}_{1}$ loss.

Finally, our specification-circuit alignment loss $\mathcal{L}_{\text{align}}$ is the sum of these three losses:
\begin{small}
\begin{equation}
    \mathcal{L}_{\text{align}} = \mathcal{L}_{\text{cont}} + \mathcal{L}_{\text{guide}} + \mathcal{L}_{\text{feat}}
\end{equation}
\end{small}%

\begin{figure}[t!]
    \centering
    \centerline{\includegraphics[width=0.7\columnwidth]{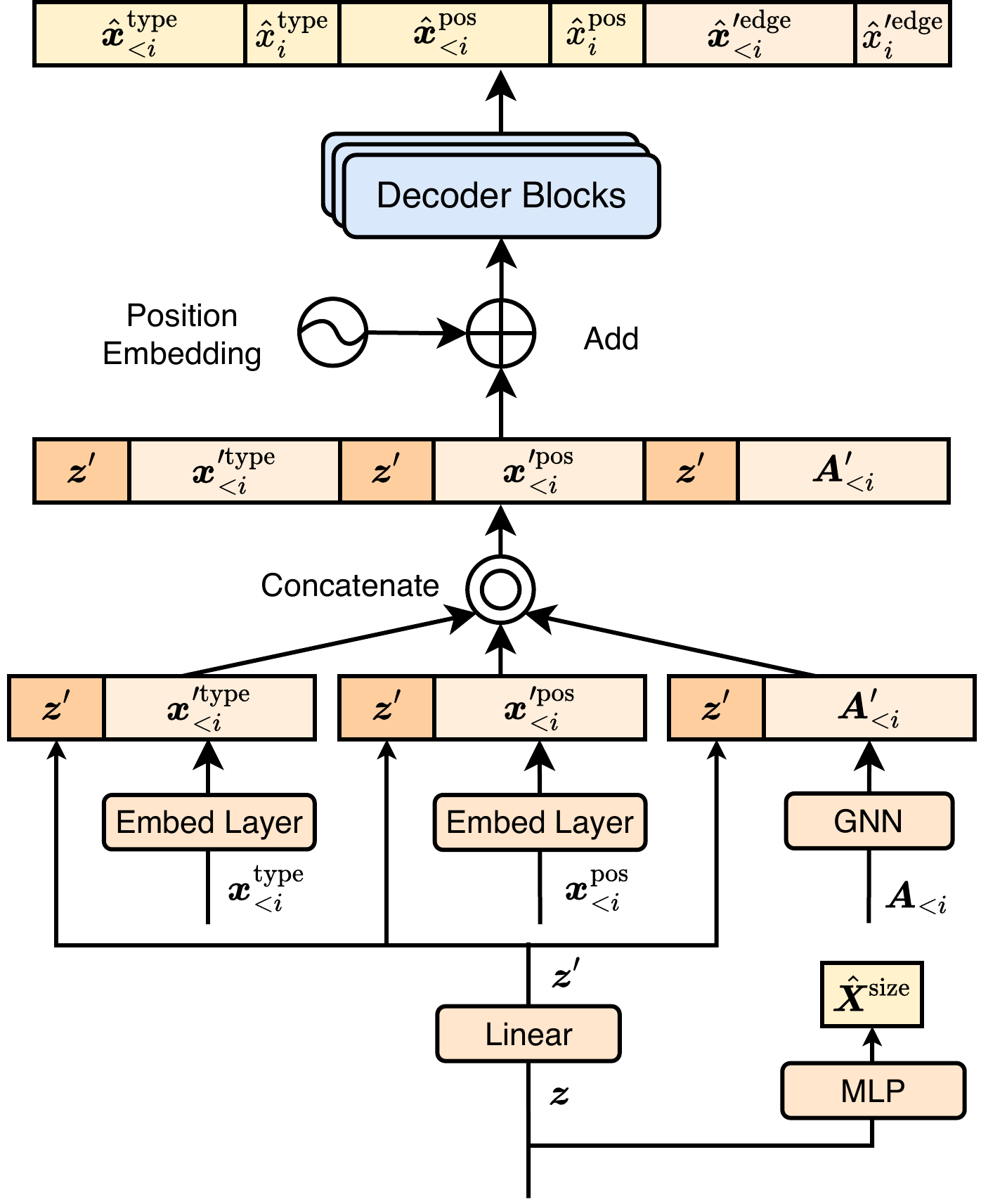}}
    \caption{
        Feature generation scheme in the circuit generator. The projected latent vector $\bm{z}'$ is concatenated with embedded node features before being processed by the transformer decoder. The definitions of the variables in the figure can be found in Section~\ref{sec:generator}.
    }
    \label{decoder}
\end{figure}

\subsubsection{Circuit generator}
\label{sec:generator}
\par We propose a generative pre-trained transformer~(GPT)-like model as the circuit generator $\mathcal{G}$. During the training stage, given a latent vector $\bm{z}$ (which can represent either $\bm{z}^{\text{ckt}}$ or $\bm{z}^{\text{spec}}$), the circuit can be reconstructed from $\bm{z}$. Initially, $\bm{z}$ is projected to $\bm{z}' \in \R^{d''}$ through a fully connected layer, where $d''$ denotes the dimension of both the projected latent vector and feature embedding in the generate phase. Then, the model generates the circuit properties sequentially, where the hat notation ($\hat{\cdot}$) denotes the generated features: the node types $\hat{\bm{x}}^{\text{type}}$, node positions $\hat{\bm{x}}^{\text{pos}}$, device parameters $\hat{\bm{X}}^{\text{size}}$, and edges $\hat{\bm{x}}^{\text{edge}}$.
\par For node types and positions, as illustrated in Fig.~\ref{decoder}, the node type vector $\bm{x}^{\text{type}}$ is first fed into the embed layer, obtaining the embedding $\bm{x}'^{\text{type}} \in \R^{N\times d''}$, then concatenated with the projected latent vector $\bm{z}'$, obtaining $\bm{x}''^{\text{type}} = [\bm{z}'; \bm{x}_1'^{\text{type}}; \bm{x}_2'^{\text{type}}; \cdots; \bm{x}_{N}'^{\text{type}}]$, where $N$ denotes the number of nodes. We then use a transformer~\cite{transformer} to decode $\bm{x}''^{\text{type}}$ to obtain $\hat{\bm{x}}^{\text{type}} = [\hat{x}^{\text{type}}_1, \hat{x}^{\text{type}}_2,  \cdots, \hat{x}^{\text{type}}_N]$, where $\hat{\bm{x}}^{\text{type}}$ represents the node type features of the generated nodes and the last node type $\hat{x}^{\text{type}}_N$ is the output, indicating the completion of the generation process. Thus, this process can be formulated in an autoregressive manner, and the output distribution of the $i$th node type can be written as the likelihood $p_{\mathcal{G}}(\hat{x}^{\text{type}}_i|\bm{x}''^{\text{type}}_{<i}, \bm{z})$. The node position features of the generated nodes $\hat{\bm{x}}^{\text{pos}}$ are obtained similarly to $\hat{\bm{x}}^{\text{type}}$. Additionally, the device parameters of the generated nodes $\hat{\bm{X}}^{\text{size}}$ are directly derived from $\bm{z}$ through an MLP-based device parameters regressor $f_{\text{size}}$, which predicts continuous sizing values for each circuit component. This conditioning mechanism enables the generator to reconstruct circuits that align with the specifications encoded in $\bm{z}$.

\par
\begin{figure}[t!]
    \centering
    \centerline{\includegraphics[width=0.5\columnwidth]{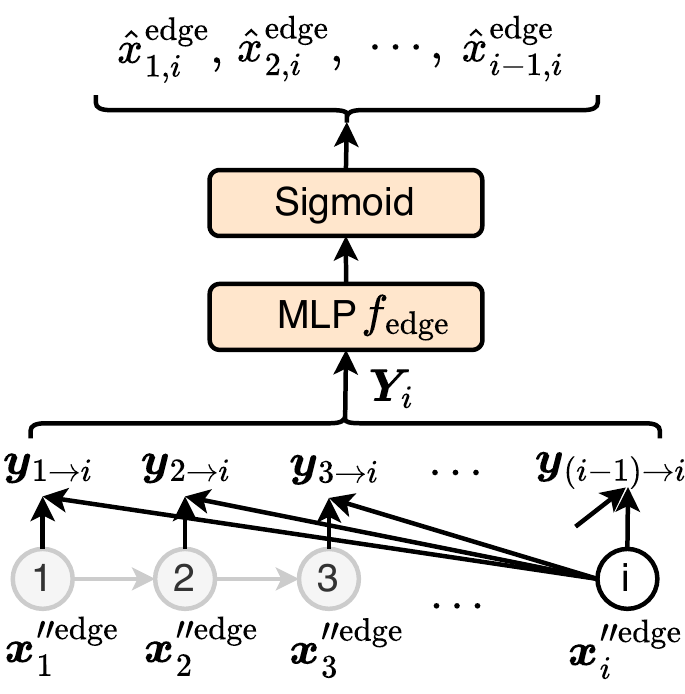}}
    \caption{
        Edge reconstruction mechanism in the circuit generator. Concatenated node embedding is processed to predict directed edge probabilities in an autoregressive manner. The definitions of the variables in the figure can be found in Section~\ref{sec:generator}.
    }
    \label{edge}
\end{figure}

 For edge generation, as shown in Fig.~\ref{decoder}, we follow a similar approach to PACE~\cite{pace}. First, the connectivity embedding $\bm{A}' \in \R^{N\times d''}$ is derived using a GNN~\cite{gnn}. Then, we concatenate $\bm{A}'$ with $\bm{z}'$ to obtain the concatenated embedding $\bm{A}''\in \R^{(N+1)\times d''}$. The transformer takes $\bm{A}''$ as input and outputs the contextualized connectivity embedding $\bm{x}'^{\text{edge}} \in \R^{(N+1)\times d''}$. We use a fully connected layer to get the projected embedding $\bm{x}''^{\text{edge}} \in \R^{(N+1)\times d}$. Next, as shown in Fig.~\ref{edge}, for the $j$th node $v_j$ and the $i$th node $v_i$, where $j<i$, we obtain the pairwise edge feature $\bm{y}_{j\rightarrow i} = [\bm{x}''^{\text{edge}}_i, \bm{x}''^{\text{edge}}_j] \in \R^{2d}$, representing the connection state of a directed edge from node $v_j$ to $v_i$. The probability of each directed edge from $v_{<i}$ to $v_i$ is computed by inputting the set of edge features representing the states of the edges connected to the $i$th node, denoted as $\bm{Y}_i = [\bm{y}_{1\rightarrow i}; \bm{y}_{2\rightarrow i}; \cdots; \bm{y}_{(i-1)\rightarrow i}] \in \R^{(i-1)\times 2d}$ into the edge prediction MLP $f_{\text{edge}}$. Formally, $f_{\text{edge}}$ is defined as a scoring function that evaluates the likelihood of a directed connection based on the pairwise edge features. The probability of an edge existing between $v_j$ and $v_i$ is defined as $p_{j \rightarrow i} = \sigma(f_{\text{edge}}(\bm{y}_{j \rightarrow i}))$, where $\sigma(\cdot)$ is the sigmoid activation function. The generated edge list $\hat{\bm{x}}^{\text{edge}}$ is obtained by applying $f_{\text{edge}}$ to the complete set of edge features $\bm{Y}=[\bm{Y}_1,\bm{Y}_2,\cdots, \bm{Y}_N] \in \R^{(N(N-1)/2)\times 2d}$, followed by the sigmoid activation function. This autoregressive design naturally preserves the DAG constraint by only considering edges from previously generated nodes.

\par During the inference stage, we first input specifications into the specification encoder, obtaining latent vectors $\bm{z}^{\text{spec}}$. Next, we feed $\bm{z}^{\text{spec}}$ into the circuit generator and generate nodes and edges in an autoregressive fashion. The decoding process stops when the model generates the node with the output type.

\paragraph{Optimization goal} 
\par We feed the encoded circuit and specification latent vectors into the circuit generator and obtain the reconstruction losses $\mathcal{L}_{\text{recon}}^{\text{ckt}}$ and $\mathcal{L}_{\text{recon}}^{\text{spec}}$, respectively. Specifically, both latent vectors are expected to reconstruct the same target circuit. The circuit reconstruction loss $\mathcal{L}_{\text{recon}}^{\text{ckt}}$ is computed using the output conditioned on $\bm{z}^{\text{ckt}}$, while $\mathcal{L}_{\text{recon}}^{\text{spec}}$ is computed using the output conditioned on $\bm{z}^{\text{spec}}$. Both losses are calculated by comparing the generated circuit features against the ground-truth values as follows:
\begin{figure*}[t!]
    \centering
    \centerline{\includegraphics[width=0.8\textwidth]{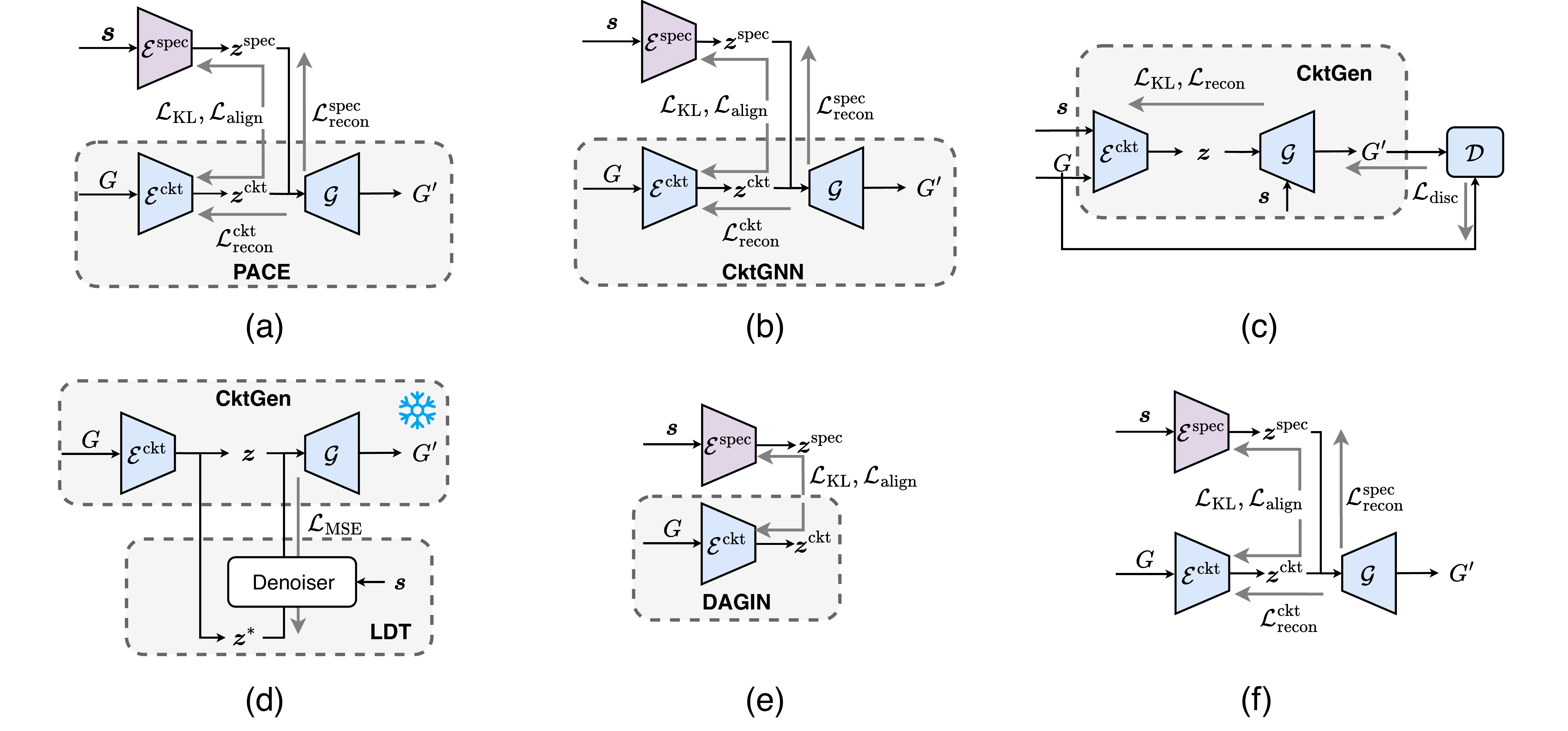}}
    \caption{Architectural comparison of related models: (a) PACE, (b) CktGNN, (c) conditional VAE generative adversarial network (CVAEGAN), (d) latent diffusion transformer (LDT), (e) surrogate, and (f) our CktGen.
        $G$: the input circuit DAG;
        $G'$: the generated circuit DAG;
        $\mathcal{D}$: discriminator;
        $\mathcal{L}_{\text{disc}}$: discriminator loss;
        $\mathcal{L}_{\text{MSE}}$: mean squared error loss;
        $\bm{z}^{*}$: latent vector with added noise.
    }
    \label{fig:arch_baseline}
\end{figure*}

\begin{small}
\begin{equation}
\begin{aligned}
\label{eq_recon_detail}
        \mathcal{L}_{\text{recon}}^{\text{ckt/spec}}=
    &\lambda_{\text{type}} \mathcal{L}_{\text{type}} (\bm{x}^{\text{type}}, \hat{\bm{x}}^{\text{type}}) +
    \lambda_{{\text{pos}}} \mathcal{L}_{\text{pos}} (\bm{x}^{\text{pos}}, \hat{\bm{x}}^{\text{pos}}) \\
    +&\mathcal{L}_{\text{edge}} (\bm{x}^{\text{edge}}, \hat{\bm{x}}^{\text{edge}}) +
    \lambda_{\text{size}}\mathcal{L}_{\text{size}} (\bm{X}^{\text{size}}, \hat{\bm{X}}^{\text{size}})
\end{aligned}
\end{equation}
\end{small}%
\noindent where $\mathcal{L}_{\text{type}}$ and $\mathcal{L}_{\text{pos}}$ represent the cross-entropy losses, $\mathcal{L}_{\text{edge}}$ denotes the binary cross-entropy loss, and $\mathcal{L}_{\text{size}}$ refers to the $\mathcal{L}_2$ loss. $\lambda_{\text{type}}$, $\lambda_{{\text{pos}}}$, and $\lambda_{\text{size}}$ are the weight hyperparameters for the node type, position, and edge reconstruction losses, respectively. $\mathcal{L}_{\text{recon}}^{\text{spec}}$ is calculated similarly to $\mathcal{L}_{\text{recon}}^{\text{ckt}}$.
The final reconstruction loss ($\mathcal{L}_{\text{recon}}$) is the total sum of these two modality specific reconstruction terms: $\mathcal{L}_{\text{recon}} = \mathcal{L}_{\text{recon}}^{\text{ckt}} + \mathcal{L}_{\text{recon}}^{\text{spec}}$.

\par In summary, our optimization goal $\mathcal{L}$ is shown as follows:
\begin{small}
\begin{equation}
    \mathcal{L}=\lambda_{\text{KL}} \mathcal{L}_{\text{KL}} + \mathcal{L}_\text{recon} + \mathcal{L}_\text{align}
\end{equation}
\end{small}%
\noindent where $\mathcal{L}$ denotes the overall optimization goal of CktGen, and $\lambda_{\text{KL}}$ is the weight of $\mathcal{L}_{\text{KL}}$.

\subsection{Baselines and surrogate}
We provide a general architectural comparison among the baselines, surrogate, and our proposed CktGen (Fig.~\ref{fig:arch_baseline}). For PACE~\cite{pace} (Fig.~\ref{fig:arch_baseline}(a)), a transformer-based DAG generative model, we adapt its architecture by incorporating a specification encoder analogous to our own. We further employ our specification-circuit alignment strategy during loss computation to ensure a fair comparison. Similarly, for CktGNN~\cite{cktgnn} (Fig.~\ref{fig:arch_baseline}(b)), a leading method for reconstruction and unconditional generation tasks, we retain the original circuit encoder and generator architectures. We augment CktGNN with a specification encoder and incorporate our specification-circuit alignment strategy in the same manner. Additionally, to conduct a more comprehensive evaluation, we implement both a simplified conditional VAE generative adversarial network~(CVAEGAN~\cite{cvaegan}; Fig.~\ref{fig:arch_baseline}(c)) and a latent diffusion transformer~(LDT~\cite{ldt}; Fig.~\ref{fig:arch_baseline}(d)), each built upon our circuit encoder and circuit generator. For CVAEGAN~\cite{cvaegan}, we introduce a discriminator $\mathcal{D}$ to compute the discriminator loss $\mathcal{L}_{\text{disc}}$. The specification condition is incorporated through an embedder that directly adds the embedded specification to the circuit latent vector. For LDT~\cite{ldt}, we employ a frozen VAE implementation, as in CktGen, to maintain architectural consistency. The denoiser is trained using the mean squared error loss $\mathcal{L}_{\text{MSE}}$. For the surrogate model (Fig.~\ref{fig:arch_baseline}(e)), we use a GIN~\cite{gin} as the circuit encoder, training only the encoder to ensure reliable evaluation. Our proposed CktGen (Fig.~\ref{fig:arch_baseline}(f)) is detailed in Section~\ref{sec:archi}.

\subsection{Implementation details}
We detail the implementation configurations for CktGen. Architecture, hyperparameters, and optimization strategies are described separately for specification-conditioned generation and reconstruction and unconditional circuit generation tasks, as settings differ to optimize performance for each task.

\paragraph{Specification-conditioned circuit generation}
For the architecture of CktGen, we employ a transformer~\cite{transformer} for both the circuit encoder and circuit generator, featuring eight attention heads and four layers. The dimension of the feed-forward layers is set to 512. The embedding and latent vector dimensions are set as follows: 128 for feature embedding in the encoding stage ($d$), 64 for latent vectors ($d'$), and 512 for feature embedding in the generative phase ($d''$).

\par Dropout rates are applied as follows: In the circuit encoder, they are set as 0.2 for the position embedding layer and GNN, and 0.3 for the transformer blocks; in the circuit generator, the dropout rate is set to 0.1. For the hyperparameters, $\lambda_{\text{type}}$ and $\lambda_{\text{pos}}$ are set at 0.5, 0.05 for Ckt-Bench-101, and at 0.7, 0.07 for Ckt-Bench-301. $\lambda_{\text{size}}$ is set at 0.01 for these two datasets. The KL loss weight $\lambda_\mathrm{KL}$ is set to $1\times 10^{-5}$. The temperature hyperparameter $\tau$ for contrastive loss is set to 0.1. 

\par During training, we used the AdamW optimizer~\cite{adamw} with a learning rate of $1\times 10^{-4}$, and the batch size was maintained at 32. All the experiments were performed on a single NVIDIA RTX 4090 graphics processing unit (GPU).

\paragraph{Reconstruction and unconditional circuit generation} 
In the reconstruction and unconditional circuit generation experiments, we set the KL loss weight to $5\times 10^{-3}$. The remaining model configuration and experimental settings were the same as those used in the specification-conditioned circuits generation.

\section{Results}
\par In this section, we first describe the datasets we used in our experiments~(Section~\ref{sec:datasets}). Then, we describe three different experiments: specification-conditioned circuit generation~(Section~\ref{sec:gen}), automated design with given target specification~(Section~\ref{sec:autodesign}), and reconstruction and unconditional circuit generation~(Section~\ref{sec:recon_uncond}). CktGen significantly outperforms existing state-of-the-art methods across all tasks and benchmarks.

\subsection{Datasets}
\label{sec:datasets}

\par We conducted experiments on the Open Circuit Benchmark~(OCB)~\cite{cktgnn}, which includes open-source analog circuits along with their specifications. OCB comprises two sub-datasets: Ckt-Bench-101 with 10000 circuits and Ckt-Bench-301 with 50000 circuits. Each dataset is divided into training and test datasets, with respective proportions of 90\% and 10\%. Each circuit joint specification class is a combination of discretized values for gain, bandwidth, and phase margin. Ckt-Bench-101 contains 376 joint specification classes, and Ckt-Bench-301 contains 238 joint specification classes.

\begin{table*}[t]
    \centering
    \begin{minipage}{\textwidth}
        \centering
        \begin{scriptsize}
        \caption{Quantitative results of specification-conditioned circuit generation on Ckt-bench-101.}
        \label{tab:cond_gen_101}
        \begin{tabular}{lccccccccc}
        \hline
        & \multicolumn{3}{c}{Retrieval precision (\%)} & & & & & \\ 
        \cline{2-4}
        
        \multirow{-1.9}{*}{Method} 
        & Top-1 
        & Top-2 
        & Top-3 
        & \multirow{-1.9}{*}{Spec-Acc (\%)} 
        & \multirow{-1.9}{*}{MM-D} 
        & \multirow{-1.9}{*}{FID} 
        & \multirow{-1.9}{*}{Inter-D}
        & \multirow{-1.9}{*}{Intra-D}
        & \multirow{-1.9}{*}{Validity (\%)} 
        \\ \hline
        PACE \cite{pace}
        & 3.415
        & 6.081
        & 8.572
        & 2.839
        & 0.684
        & \underline{5.653}
        & 6.964
        & 6.064
        & 64.16
        \\
        CktGNN \cite{cktgnn}
        & 1.160
        & 2.141
        & 3.078
        & 1.000
        & 0.964
        & 33.92
        & 5.907
        & 5.521
        & 85.28
        \\
        CVAEGAN
        & 0.789
        & 1.468
        & 2.179
        & 0.635
        & 0.865
        & \textbf{4.782}
        & 7.575
        & 7.362
        & 98.05
        \\
        LDT
        & 2.371
        & 4.981
        & 7.471
        & 1.915
        & 0.598
        & 30.72
        & \textbf{11.26}
        & 7.200
        & \underline{98.32}
        \\
        \rowcolor{aliceblue}
        CktGen
        & \textbf{35.73}
        & \textbf{55.93}
        & \textbf{65.21}
        & \textbf{47.57}
        & \textbf{0.385}
        & 6.092
        & \underline{8.574}
        & 1.987
        & 95.47
        \\
        w/o $\mathcal{L}_{\text{align}}$
        & 1.965
        & 3.883
        & 5.424
        & 1.638
        & 0.725
        & 6.765
        & 7.444
        & 6.579
        & \textbf{99.12}
        \\
        w/o $\mathcal{L}_{\mathrm{KL}}$
        & \underline{32.45}
        & \underline{52.07}
        & \underline{62.64}
        & \underline{43.39}
        & \underline{0.390}
        & 6.574
        & 8.486
        & 0.000
        & 95.84
        \\
        w/o $\mathcal{L}_{\mathrm{KL}}$ and $\mathcal{L}_{\text{align}}$
        & 0.000
        & 0.000
        & 0.754
        & 0.377
        & 1.041
        & 33.58
        & 6.014
        & 0.000
        & 96.22
        \\ \hline
        \end{tabular}
        \end{scriptsize}

        \vspace{3pt}
        \begin{footnotesize}
        \textbf{Bold} indicates the best results, \underline{underline} indicates the second best, and w/o denotes the absence of a specific loss term. Retrieval precision, Spec-Acc, and validity are higher-is-better metrics, while MM-D and FID are lower-is-better metrics. For diversity metrics, Inter-D measures the distance between circuits generated under different joint specification classes, while Intra-D represents the distance among circuits generated for the same joint specification class. The key requirement is that Inter-D should be greater than Intra-D, indicating effective clustering of circuits by joint specification classes. Under this requirement, higher values for both Inter-D and Intra-D are desirable, as they indicate greater diversity across and within specification classes, respectively.
        \end{footnotesize}
    \end{minipage}
\end{table*}

\begin{table*}[t]
    \centering
    \begin{scriptsize}
    \caption{Quantitative results of specification-conditioned circuit generation on Ckt-bench-301.}
    \label{tab:cond_gen_301}
    \begin{tabular}{lccccccccc}
        \hline
        & \multicolumn{3}{c}{Retrieval precision (\%)} & & & & & \\ 
        \cline{2-4}
        
        \multirow{-1.9}{*}{Method} 
        & Top-1 
        & Top-2 
        & Top-3 
        & \multirow{-1.9}{*}{Spec-Acc (\%)} 
        & \multirow{-1.9}{*}{MM-D} 
        & \multirow{-1.9}{*}{FID} 
        & \multirow{-1.9}{*}{Inter-D}
        & \multirow{-1.9}{*}{Intra-D}
        & \multirow{-1.9}{*}{Validity (\%)} 
        \\ \hline
        PACE \cite{pace}
        & 4.188
        & 8.263
        & 12.16
        & 3.649
        & 0.485
        & 4.795
        & 5.714
        & 4.158
        & 69.39
        \\
        CktGNN \cite{cktgnn}
        & 2.059
        & 3.825
        & 5.723
        & 1.321
        & 0.862
        & 27.65
        & 4.761
        & 4.032
        & 85.56
        \\
        CVAEGAN
        & 0.796
        & 1.479
        & 2.268
        & 0.730
        & 0.815
        & 6.311
        & 6.033
        & 5.801
        & 95.93
        \\
        LDT
        & 4.313
        & 8.171
        & 11.87
        & 4.104
        & 0.378
        & 7.016
        & \textbf{8.474}
        & 5.343
        & \underline{99.16}
        \\
        \rowcolor{aliceblue}
        CktGen
        & \textbf{26.81}
        & \textbf{42.16}
        & \textbf{52.42}
        & \underline{22.64}
        & \textbf{0.269}
        & \textbf{2.136}
        & 6.875
        & 0.988
        & 97.99
        \\   
        w/o $\mathcal{L}_{\text{align}}$
        & 2.041
        & 4.225
        & 6.284
        & 2.000
        & 0.568
        & 6.390
        & 6.234
        & 4.721
        & 97.22
        \\
        w/o $\mathcal{L}_{\mathrm{KL}}$
        & \underline{24.66}
        & \underline{39.64}
        & \underline{48.89}
        & \textbf{23.34}
        & \underline{0.277}
        & \underline{2.180}
        & \underline{6.922}
        & 0.000
        & \textbf{99.55}
        \\
        w/o $\mathcal{L}_{\mathrm{KL}}$ and $\mathcal{L}_{\text{align}}$
        & 0.000
        & 1.321
        & 2.643
        & 0.000
        & 1.165
        & 41.85
        & 4.263
        & 0.000
        & 92.07
        \\ \hline
    \end{tabular}
    \end{scriptsize}
\end{table*}
\begin{table*}[t!]
    \centering
    \begin{scriptsize}
        \caption{Ablation study of $\mathcal{L}_{\mathrm{KL}}$ and $\tau$ for specification-conditioned circuit generation on Ckt-Bench-101.}
        \label{tab:abl_kl_tau_gen}
        \begin{tabular}{llccccccccc}
        \hline
        & & \multicolumn{3}{c}{Retrieval precision (\%)} & & & & & \\ 
        \cline{3-5}
        
        \multirow{-1.9}{*}{$\lambda_{\text{KL}}$} 
        & \multirow{-1.9}{*}{$\tau$}
        & Top-1 
        & Top-2 
        & Top-3 
        & \multirow{-1.9}{*}{Spec-Acc (\%)} 
        & \multirow{-1.9}{*}{MM-D} 
        & \multirow{-1.9}{*}{FID} 
        & \multirow{-1.9}{*}{Inter-D}
        & \multirow{-1.9}{*}{Intra-D}
        & \multirow{-1.9}{*}{Validity (\%)} 
        \\ \hline

        $10^{-7}$
        & 0.100
        & 34.58
        & 52.63
        & 62.64
        & 40.47
        & 0.393
        & 6.637
        & 8.520
        & 0.906
        & 94.00
        \\
        
        $10^{-6}$
        & 0.100
        & 33.66
        & 50.51
        & 60.90
        & 40.74
        & 0.401
        & 6.387
        & 8.519
        & 0.999
        & \underline{95.74}
        \\
        
        \rowcolor{aliceblue}
        $10^{-5}$
        & 0.100
        & \textbf{35.73}
        & 55.93
        & \textbf{65.21}
        & \textbf{47.57}
        & \underline{0.385}
        & 6.092
        & 8.574
        & 1.987
        & 95.47
        \\
        
        $10^{-4}$
        & 0.100
        & \underline{35.29}
        & \underline{56.01}
        & 63.11
        & 45.44
        & 0.393
        & 5.867
        & 8.602
        & 2.472
        & \textbf{96.74}
        \\
        
        $10^{-3}$
        & 0.100
        & 33.42
        & 53.87
        & 61.39
        & 43.18
        & 0.402
        & \textbf{5.359}
        & 8.576
        & 3.166
        & 94.38
        \\ 
        
        $10^{-2}$
        & 0.100
        & 23.63
        & 38.34
        & 45.40
        & 32.46
        & 0.454
        & 5.766
        & \textbf{8.701}
        & 4.539
        & 94.68
        \\
        
        $10^{-5}$
        & 0.001
        & 33.29
        & 54.20
        & 63.65
        & \underline{47.03}
        & \textbf{0.383}
        & \underline{5.438}
        & \underline{8.614}
        & 1.914
        & 93.76
        \\
        
        $10^{-5}$
        & 0.010
        & 34.83
        & 52.77
        & 59.50
        & 42.11
        & 0.410
        & 5.956
        & 8.596
        & 1.813
        & 94.13
        \\
        
        \rowcolor{aliceblue}
        $10^{-5}$
        & 0.100
        & \textbf{35.73}
        & 55.93
        & \textbf{65.21}
        & \textbf{47.57}
        & \underline{0.385}
        & 6.092
        & 8.574
        & 1.987
        & 95.47
        \\
        
        $10^{-5}$
        & 1.000
        & 34.87
        & \textbf{56.73}
        & \underline{64.76}
        & 44.70
        & 0.395
        & 6.359
        & 8.612
        & 1.638
        & 94.38
        \\
          
        \hline
        \end{tabular}
        \end{scriptsize}
\end{table*}

\subsection{Specification-conditioned circuit generation}
\label{sec:gen}
\par Our framework takes interval-based joint specification classes as input and requires corresponding circuits to be generated based on these specifications. To evaluate specification-to-circuit generation performance, we first group the circuits in the test dataset by joint specification class. For each joint specification class, we sample a specification latent vector using the reparameterization trick~\cite{vae} and randomly select a ground-truth circuit. We then decode this specification latent to a circuit. Finally, using CktGen as a pre-trained surrogate model, we encode both the generated circuit and its specification condition into the latent vectors. 

\begin{figure*}[t!]
    \centering
    \centerline{\includegraphics[width=0.7\textwidth]{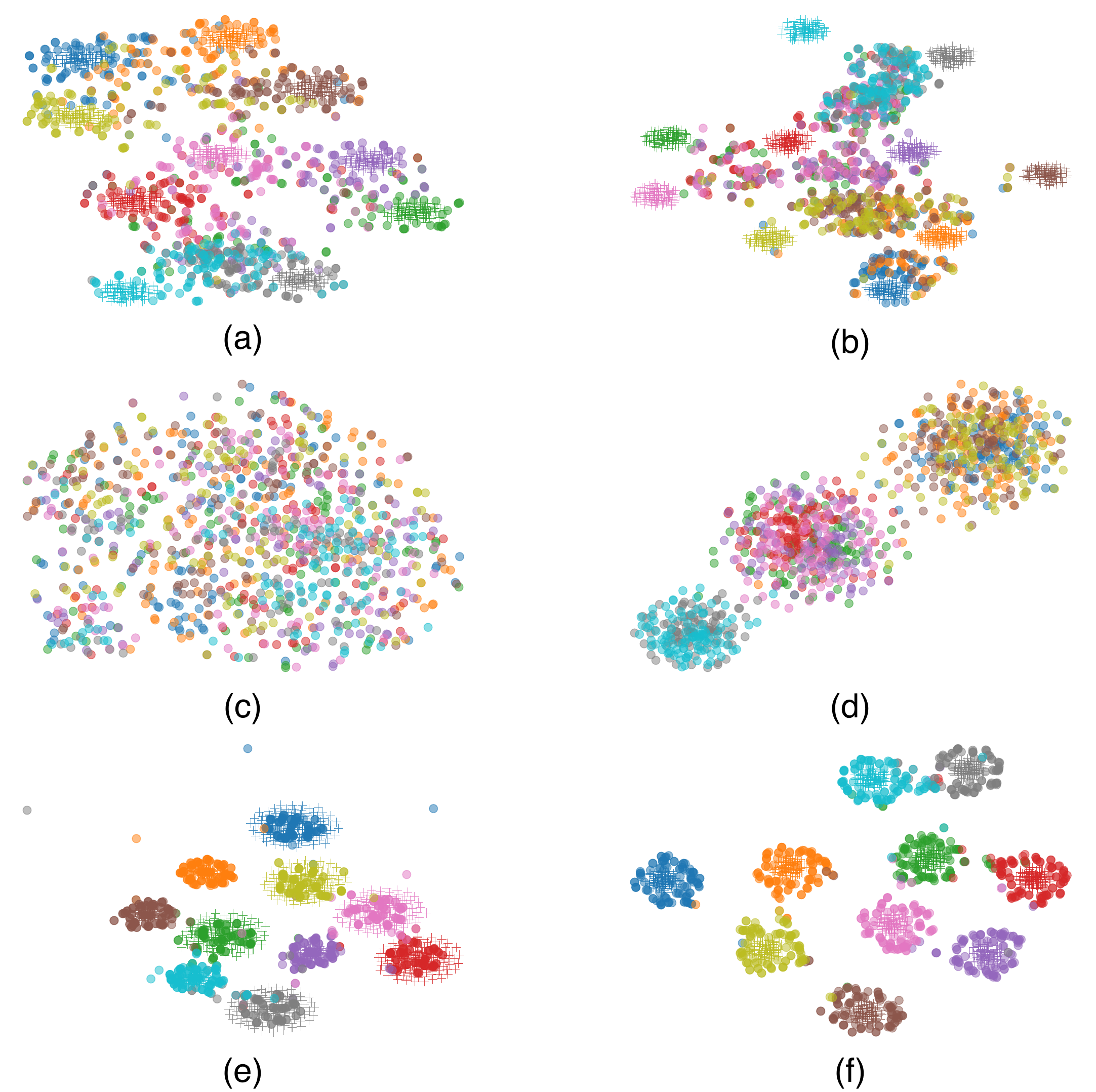}}
    \caption{
        \textit{t}-SNE visualization of latent representations for different models on Ckt-Bench-101: (a) PACE, (b) CktGNN, (c) CVAEGAN, (d) LDT, (e) surrogate, and (f) our CktGen. Circuit latents are shown as circles and specification latents as ``+'' symbols, with each color representing a different joint specification class.
    }
    \label{fig:tsne}
  \end{figure*}

\par We evaluate our method using the following metrics for 120 rounds on the test dataset. Consistency metrics: \ding{182} Retrieval precision (top-\textit{K}) measures the proportion of generated circuits relevant to the specification query based on cosine similarity in the latent space. \ding{183} Specification accuracy (Spec-Acc) quantifies the proportion of generated circuits where the specifications estimated by the surrogate model match the original generation conditions. \ding{184} Multimodal distance (MM-D)~\cite{t2m_gpt} calculates the average cosine distance between encoded latent vectors of the generated circuits and specifications. \ding{185} Frechet Inception Distance (FID)~\cite{fid} evaluates distributional differences between the generated circuits and the ground truth. Diversity metrics: \ding{186} Inter-class diversity (Inter-D) measures the average Euclidean distance between circuits generated from different joint specification classes. \ding{187} Intra-class diversity (Intra-D) measures the average Euclidean distance between circuits generated from the same joint specification class. \ding{188} Validity measures the proportion of generated circuits that are valid DAGs with a single input and output node, free of cycles, and without feedback paths in the main path~\cite{cktgnn}. More detailed formulations of the evaluation metrics are shown in \ref{appd:sec:c}. Quantitatively, we compare CktGen with baseline models on specification-conditioned generation. Qualitatively, we analyze the latent encodings of circuits and specifications from the dataset to demonstrate the model's one-to-many mapping capability.

\subsubsection{Quantitative results}
\par We conducted the specification-conditioned analog circuit generation experiments on Ckt-Bench-101 (Table~\ref{tab:cond_gen_101}~\cite{cktgnn, pace}) and Ckt-Bench-301 (Table~\ref{tab:cond_gen_301}~\cite{cktgnn, pace}). CktGen consistently outperforms all baseline methods across both datasets. The key challenge in specification-conditioned generation is learning the one-to-many mapping from specifications to valid circuit implementations. CktGen addresses this challenge effectively, achieving top-1 retrieval precision of 35.73\% on Ckt-Bench-101, representing more than ten-fold improvements over existing methods. This indicates that CktGen can reliably identify and generate circuits corresponding to given specifications, a capability essential for practical design automation. More importantly, CktGen achieves Spec-Acc of 47.57\% on Ckt-Bench-101, substantially exceeding all baselines (below 3\%). This metric directly measures whether generated circuits meet target specifications, and the substantial gap demonstrates that CktGen's specification-circuit alignment strategy is critical for learning accurate conditional generation. CktGen achieves the lowest MM-D values and competitive FID scores, indicating that generated circuits are not only functionally valid but also closely aligned with target specifications in the latent space. For diversity metrics, CktGen exhibits Inter-D values that exceed Intra-D values on both datasets, demonstrating successful clustering by joint specification class while maintaining diversity within each class. This pattern indicates that CktGen learns a well-structured latent space where circuits with similar specifications are grouped together, yet can still generate multiple distinct implementations for the same specification, enabling design space exploration. CktGen maintains high validity rates (above 95\%) on both datasets, ensuring that generated circuits are structurally sound and can be directly used in design workflows.

\par The baseline methods reveal distinct failure modes. CVAEGAN achieves a competitive FID on Ckt-Bench-101, suggesting it can generate circuits with realistic distributions, but performs poorly on retrieval precision and Spec-Acc (below 1\%). This indicates that while CVAEGAN can generate diverse and valid circuits, it fails to establish the critical link between specifications and circuit implementations. LDT achieves high Inter-D values and validity, demonstrating its ability to generate diverse and structurally valid circuits. However, its weak performance in retrieval precision and Spec-Acc reveals that the generated diversity does not align with specification requirements, suggesting that the diffusion process does not effectively incorporate specification constraints. PACE and CktGNN yield uniformly low scores across all metrics, with top-1 retrieval precision and Spec-Acc both below 5\%. These results demonstrate that even when equipped with the alignment strategy employed by CktGen, existing architectures remain incapable of capturing the complex one-to-many mapping relationships. In contrast, CktGen uniquely integrates high specification alignment, circuit validity, and controlled diversity, enabling effective design space exploration while ensuring strict adherence to target requirements. 

\begin{figure*}[t!]
    \centering
    \centerline{\includegraphics[width=0.7\textwidth]{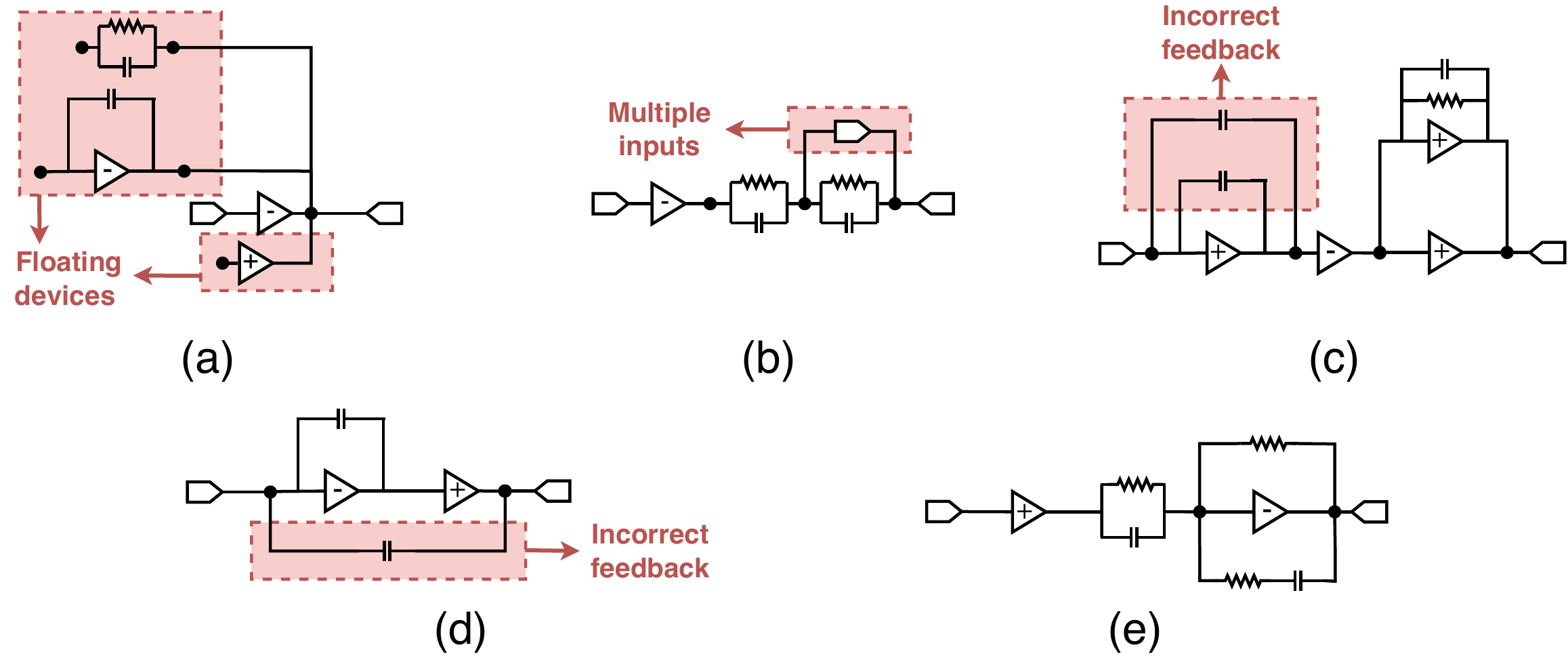}}
    \caption{
        Circuits generated by different models under given specifications~(gain=2, bandwidth=15, and PM=2): (a) PACE, (b) CktGNN, (c) CVAEGAN, (d) LDT, and (e) our CktGen. Red dashed boxes highlight electrical errors in baseline models.
    }
    \label{fig:gen_ckts}
\end{figure*}
  
\begin{figure}[t!]
  \centering
  \centerline{\includegraphics[width=0.8\columnwidth]{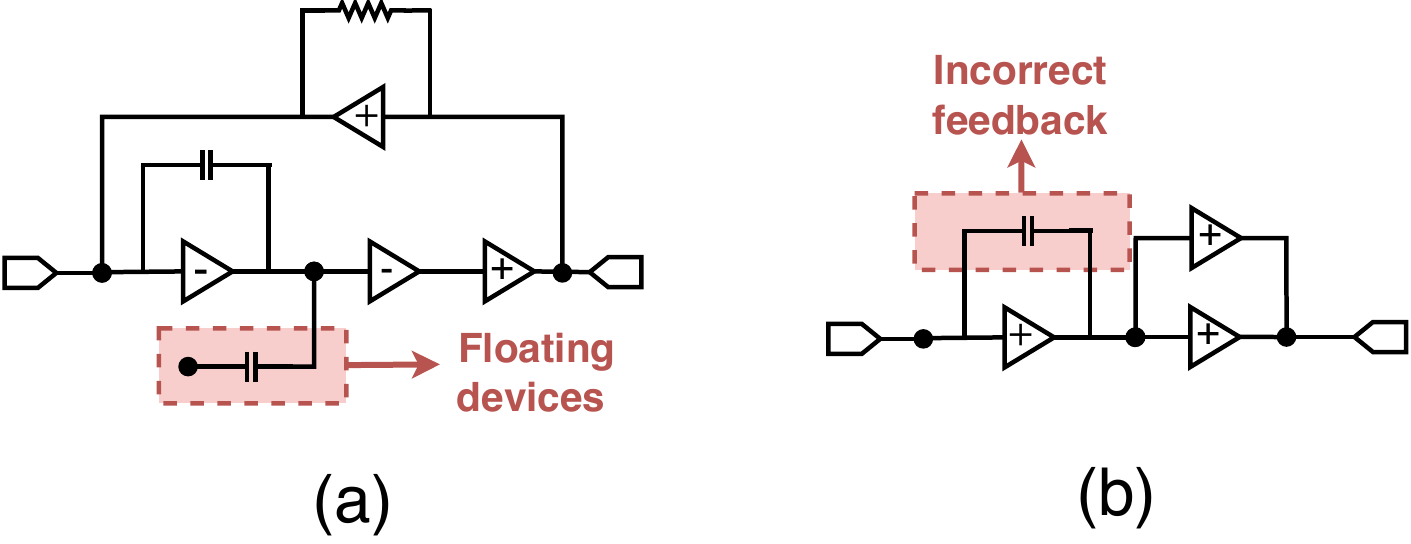}}
  \caption{
      Failure cases of CktGen for underrepresented specifications in the training data: (a) gain=2, bandwidth=22, and PM=1; (b) gain=2, bandwidth=3, and PM=1.
  }
  \label{fig:fail_case}
\end{figure}

\par To better understand the impact of hyperparameters and components in CktGen on the specification-conditioned circuit generation task, we first conducted an ablation study of its key components, which indicated that the proposed specification-circuit alignment loss $\mathcal{L}_{\text{align}}$ is the key to the performance improvement. Since PACE and CktGNN also employ this alignment loss during training but are poor in performance, their inferior results suggest that their VAE architectures lack the capacity to effectively learn this challenging task. When trained without the KL loss ($\mathcal{L}_{\mathrm{KL}}$), CktGen degenerates into a deterministic generator that yields a single circuit for each joint specification class, collapsing into a one-to-one mapping and exhibiting 0 Intra-D. When both $\mathcal{L}_{\mathrm{KL}}$ and $\mathcal{L}_{\text{align}}$ are removed, CktGen fails to generate specification-aligned circuits, with Top-1 retrieval precision dropping to near zero and Spec-Acc falling below 0.4\%, indicating that both losses are essential for learning the specification-conditioned generation task. Next, we performed an ablation study on two critical hyperparameters: the KL loss weight $\lambda_\mathrm{KL}$ and the temperature hyperparameter $\tau$ in contrastive training. The results are shown in Table~\ref{tab:abl_kl_tau_gen}. For $\lambda_\mathrm{KL}$, we explored values ranging from $10^{-7}$ to $10^{-2}$. Since our goal was to generate high-quality circuits with given specifications, we fixed the $\lambda_\mathrm{KL}$ at $10^{-5}$. Similarly, for the temperature hyperparameter $\tau$, we evaluated values from $0.001$ to $1$ and selected the optimal setting. Additionally, we investigated the impact of training epochs on model performance for CktGen, CVAEGAN, and LDT; the results are provided in \ref{appd:sec:d}.

\begin{figure*}[t!]
    \centering
    \centerline{\includegraphics[width=0.8\textwidth]{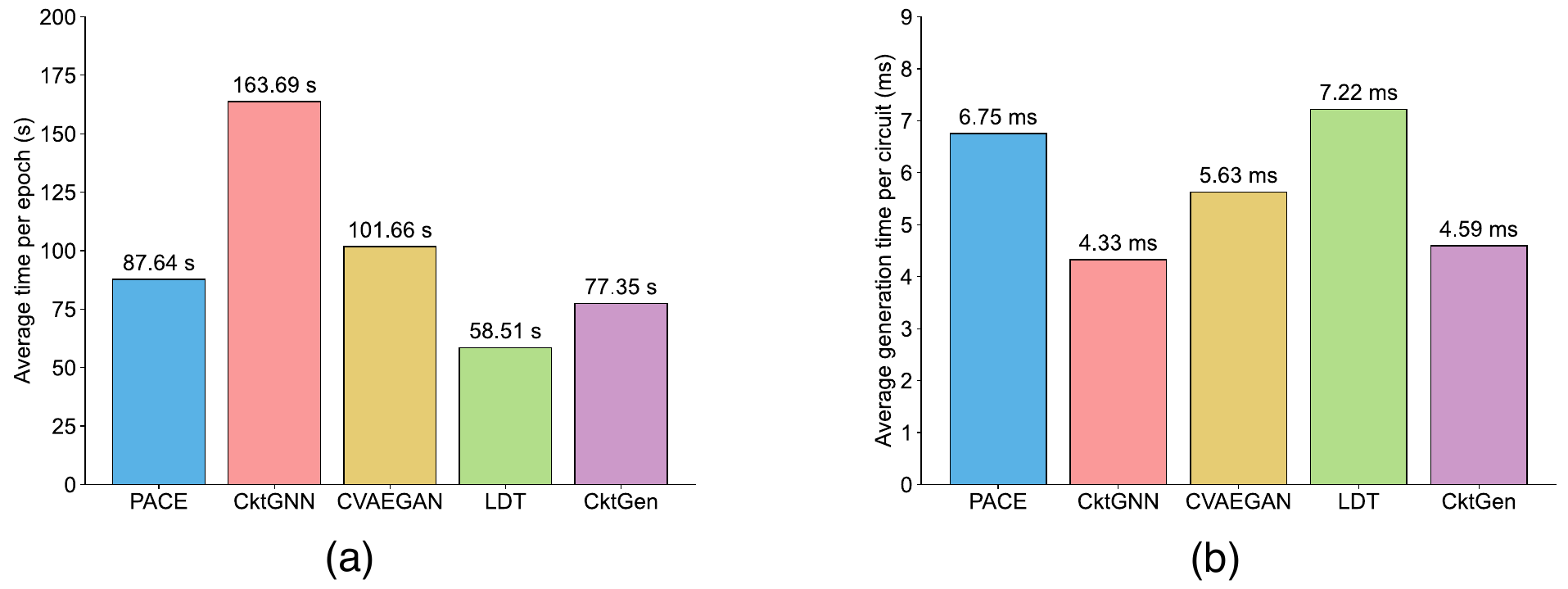}}
    \caption{Comparison of (a) training and (b) inference efficiency.}
    \label{fig:cond_gen_infer}
\end{figure*}

\subsubsection{Qualitative results}
We provide qualitative analysis through two complementary perspectives: latent space clustering visualization to assess the discriminative capability of learned representations, and direct inspection of generated circuit topologies to evaluate generation quality and validity.

\paragraph{Clustering capability} To assess the clustering capability of our models against the baselines, we randomly selected 100 ground-truth circuits for each of 10 joint specification classes from the Ckt-Bench-101 dataset and visualized the latent representations using \textit{t}-distributed stochastic neighbor embedding~(\textit{t}-SNE). For models without a specification encoder (\ie, CVAEGAN and LDT), only the circuit latents are shown. For other models, circuit latents are depicted as circles and specification latents as ``+'' symbols, with each color representing a joint specification class. As illustrated in Fig.~\ref{fig:tsne}, PACE (Fig.~\ref{fig:tsne}(a)) and CktGNN (Fig.~\ref{fig:tsne}(b)) exhibit multiple clusters but with less compactness and significant overlap, indicating weaker disentanglement between joint specification classes. CVAEGAN shows a complete lack of clustering (Fig.~\ref{fig:tsne}(c)), with all data points forming a single amorphous cloud without any discernible separation, demonstrating its inability to distinguish between different joint specification classes. LDT displays several distinct clusters (Fig.~\ref{fig:tsne}(d)), but they are more elongated and less uniformly compact, suggesting suboptimal organization of the latent space. The surrogate model achieves very compact and exceptionally well-separated clusters (Fig.~\ref{fig:tsne}(e)), with tight grouping that demonstrates effective class discrimination. CktGen achieves clear, compact, and well-separated clusters comparable to the surrogate (Fig.~\ref{fig:tsne}(f)), demonstrating superior capability to distinguish and organize circuits by joint specification class. These results highlight that while some baseline methods can form clusters, CktGen uniquely achieves both compact clustering and clear separation, enabling effective organization of the latent space for specification-conditioned generation.

\paragraph{Qualitative results of conditional generation} To provide qualitative insights into model generation capability, we visualized circuits generated by CktGen and baseline models under identical target specifications (gain=2, bandwidth=15, and PM=2) in Fig.~\ref{fig:gen_ckts}. Across PACE (Fig.~\ref{fig:gen_ckts}(a)), CktGNN~(Fig.~\ref{fig:gen_ckts}(b)), CVAEGAN~(Fig.~\ref{fig:gen_ckts}(c)), and LDT~(Fig.~\ref{fig:gen_ckts}(d)), the baselines frequently yield structurally invalid topologies, including floating nodes, multi-input node conflicts, and incorrect feedback loops. These errors, highlighted by red dashed boxes in Fig.~\ref{fig:gen_ckts}, violate fundamental analog design constraints and render the circuits non-functional. By contrast, CktGen~(Fig.~\ref{fig:gen_ckts}(e)) produces topologies that satisfy key validity checks, indicating that it learns and enforces physical design constraints. We also report failure cases for CktGen under input specifications that are relatively underrepresented in the training data, as shown in Fig.~\ref{fig:fail_case}. Although infrequent, these include floating nodes and incorrect feedback paths, pointing to opportunities for further refinement.

\subsubsection{Training and inference efficiency}

\par We evaluated the training and inference efficiency of CktGen compared with those of the baseline models on the specification-conditioned circuit generation task (Fig.~\ref{fig:cond_gen_infer}(a) and (b)). For the LDT model, the reported training time is the sum of the VAE and denoiser stages. In terms of training efficiency (Fig.~\ref{fig:cond_gen_infer}(a)), PACE requires 87.64~s per epoch, while CktGNN exhibits the highest training cost at 163.69~s per epoch, nearly double that of PACE. CVAEGAN requires 101.66~s per epoch, and LDT achieves the lowest training time at 58.51~s per epoch. CktGen achieves 77.35~s per epoch, representing a 53\% reduction compared with CktGNN and remaining competitive with other baselines. For inference efficiency (Fig.~\ref{fig:cond_gen_infer}(b)), PACE requires 6.75~ms per circuit, while CktGNN achieves the fastest inference at 4.33~ms per circuit. CVAEGAN and LDT require 5.63~ms and 7.22~ms per circuit, respectively, with LDT exhibiting the highest inference latency. CktGen achieves 4.59~ms per circuit, closely matching CktGNN's inference speed and representing a 36\% improvement over LDT. All results are averaged across Ckt-Bench-101 and Ckt-Bench-301. These findings demonstrate that CktGen achieves a favorable balance between training and inference efficiency, delivering fast inference comparable to the best-performing baseline while maintaining competitive training costs.

\begin{figure*}[t!]
    \centering
    \centerline{\includegraphics[width=0.8\textwidth]{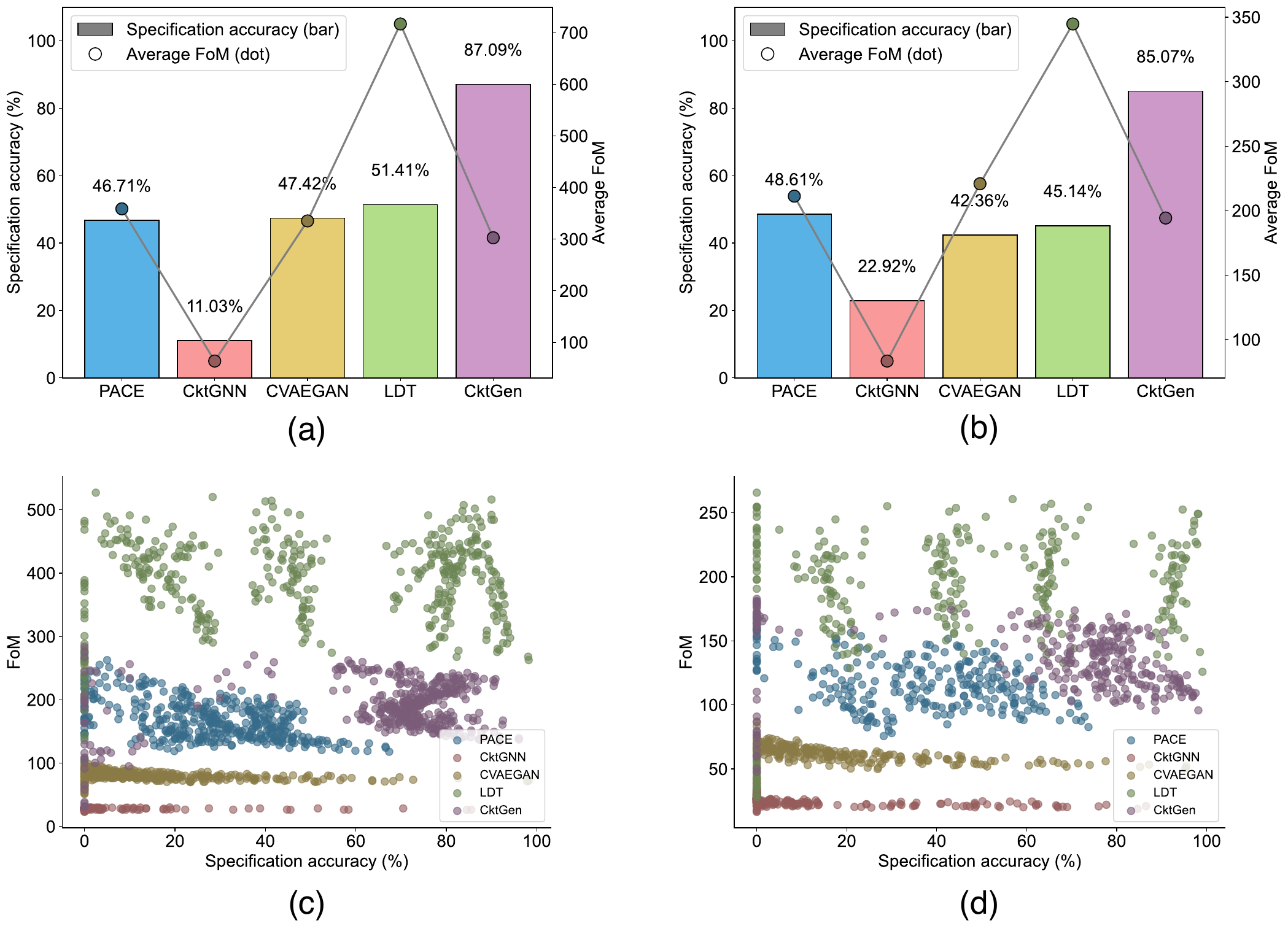}}
    \caption{
        Automated analog circuit design with target specifications. 
        (a, b): Spec-Acc (bar) and average FoM (dot) of the best-found circuits for each target specification set across different methods on (a) Ckt-Bench-101 and (b) Ckt-Bench-301. 
        (c, d): The distribution of Spec-Acc and FoM of the generated circuits at each sampling step during MAB optimization: (c) Ckt-Bench-101 and (d) Ckt-Bench-301.
    }
    \label{fig:auto_design}
\end{figure*}

\subsection{Automated design with given target specification}
\label{sec:autodesign}
\par Our trained model is designed to generate valid circuits with a target joint specification class. For a target such as $s_{\text{Gain}} > s^{*}_{\text{Gain}}$, $s_{\text{BW}} > s^{*}_{\text{BW}}$, and $s_{\text{PM}} > s^{*}_{\text{PM}}$, any joint specification class that satisfies these constraints and exists in the datasets is considered valid. The FoM metric quantifies the overall trade-off among circuit specifications. To identify the optimal circuit, we employ a MAB algorithm to search for the highest FoM design among all valid candidates.

\par To evaluate model performance in this realistic setting, we use all the joint specification classes in the datasets as target specifications. Furthermore, to evaluate the robustness, we also randomly sample 50 target joint specification classes, ensuring that each has valid candidate circuits in the datasets. For each method, we use a surrogate model to evaluate the generated circuit's specifications and its FoM value, and apply the same multi-armed bandit algorithm for fair comparison across CktGen and baseline models.

\par As shown in Fig.~\ref{fig:auto_design}(a) and (b), we report both the average FoM and the Spec-Acc. Across Ckt-Bench-101 and Ckt-Bench-301, the pattern is consistent: PACE attains moderate Spec-Acc (46.71\% and 48.61\%), while CktGNN attains low Spec-Acc (11.03\% and 22.92\%) with the poorest FoM. CVAEGAN and LDT improve to moderate Spec-Acc (around 40\%--50\%) while achieving higher FoM, with LDT typically leading FoM among baselines. CktGen delivers the highest Spec-Acc on both datasets (87.09\% and 85.07\%) while retaining competitive FoM, indicating the best balance between meeting target specifications and circuit quality. This level of Spec-Acc is critical for ensuring that generated designs satisfy target requirements in practical design tasks.

\par Fig.~\ref{fig:auto_design}(c) and (d) show a similar distributional structure on both datasets during MAB optimization: CktGNN clusters at low Spec-Acc and low FoM; PACE spans a broader low-to-mid Spec-Acc region with competitive FoM; CVAEGAN and LDT occupy the mid-accuracy band (40\%--60\%) with higher FoM, with LDT showing slightly higher FoM concentration; CktGen remains tightly clustered at high Spec-Acc (80\%--100\%) with stable FoM. These consistent patterns highlight that CktGen uniquely achieves both high specification satisfaction and reliable FoM, whereas baselines either remain at lower accuracy levels or trade accuracy for higher FoM.

\begin{table*}[t!]
    \centering
    \begin{scriptsize}
        \caption{Comparison of reconstruction and unconditional generation on Ckt-Bench-101 and Ckt-Bench-301. Higher is better for all metrics.}
        \label{exp_recon_uncond}
        \begin{tabular}{lcccccc}
        \hline
            & \multicolumn{3}{c}{Ckt-Bench-101}
            & \multicolumn{3}{c}{Ckt-Bench-301} \\
            \cline{2-7}
            
            \multirow{-1.9}{*}{Methods}
            & Recon-Acc (\%)
            & Validity (\%)
            & Novelty (\%)
            & Recon-Acc (\%)
            & Validity (\%) 
            & Novelty(\%) \\ \hline
            
            CktGNN~\cite{cktgnn} 
            & 45.3 
            & \underline{91.9} 
            & \underline{96.3}
            & \underline{99.0} 
            & \underline{97.8} 
            & \underline{95.2}\\
    
            PACE \cite{pace} 
            & \underline{99.7} 
            & 72.9 
            & \textbf{96.7 }
            & \textbf{99.9} 
            & 80.5 
            & \textbf{97.7} \\
    
            \rowcolor{aliceblue}
            CktGen
            & \textbf{99.9} 
            & \textbf{98.7}
            & 95.0 
            & \textbf{99.9} 
            & \textbf{98.4} 
            & 93.1 \\
    
        \hline
        \end{tabular}
    \end{scriptsize}
\end{table*}

\begin{table*}[t]
    \centering
    \begin{minipage}{0.48\textwidth}
        \centering
        \begin{scriptsize}
        \caption{Ablation study of $\lambda_{\text{type}}$ and $\lambda_{\text{pos}}$ for reconstruction and unconditional generation on Ckt-Bench-101.}
        \label{tab:abl_recon_101}
        \begin{tabular}{llccc}
        \hline
            $\lambda_{\text{type}}$ 
            & $\lambda_{\text{pos}}$ 
            & Recon-Acc (\%) 
            & Validity (\%) 
            & Novelty (\%) \\ \hline

            0.1 & 0.01 & 26.0 & 96.6 & 88.3 \\
            0.2 & 0.02 & 80.2 & 98.1 & 89.5 \\
            0.3 & 0.03 & 85.5 & 97.8 & 91.4 \\
            0.4 & 0.04 & 98.1 & 96.4 & 93.4 \\
            \rowcolor{aliceblue}
            0.5 & 0.05 & \textbf{99.9} & \textbf{98.7 } & \textbf{94.9} \\
            0.6 & 0.06 & 97.6 & \underline{98.2} & \underline{94.6} \\
            0.7 & 0.07 & 96.7 & 96.8 & 93.8 \\
            0.8 & 0.08 & 98.2 & 97.3 & 92.4 \\
            0.9 & 0.09 & \underline{99.8} & 95.8 & 92.8 \\
            1.0 & 0.10 & 98.8 & 97.0 & 93.4 \\
            1.0 & 1.00 & 99.0 & 96.7 & 92.4 \\ 
        \hline
        \end{tabular}
        \end{scriptsize}
    \end{minipage}
    \hfill
    \begin{minipage}{0.48\textwidth}
        \centering
        \begin{scriptsize}
        \caption{Ablation study of $\lambda_{\text{type}}$ and $\lambda_{\text{pos}}$ for reconstruction and unconditional generation on Ckt-Bench-301.}
        \label{tab:abl_recon_301}
        \begin{tabular}{llccc}
        \hline
            $\lambda_{\text{type}}$ 
            & $\lambda_{\text{pos}}$ 
            & Recon-Acc (\%)
            & Validity (\%)
            & Novelty (\%) \\ \hline

            0.1 & 0.01 & 98.3 & 97.9 & 87.6 \\
            0.2 & 0.02 & 99.6 & \underline{98.2} & 89.1 \\
            0.3 & 0.03 & \underline{99.7} & 97.8 & 92.1 \\
            0.4 & 0.04 & \textbf{99.9} & 96.9 & 91.9 \\
            0.5 & 0.05 & \textbf{99.9} & 96.9 & 92.7 \\
            0.6 & 0.06 & \textbf{99.9} & 97.3 & \underline{93.1} \\
            \rowcolor{aliceblue}
            0.7 & 0.07 & \textbf{99.9} & \textbf{98.4} & \underline{93.1} \\
            0.8 & 0.08 & \textbf{99.9} & 95.5 & 92.3 \\
            0.9 & 0.09 & \textbf{99.9} & 97.4 & 92.6 \\
            1.0 & 0.10 & \textbf{99.9} & 97.9 & 92.6 \\
            1.0 & 1.00 & \textbf{99.9} & 97.2 & \textbf{93.7} \\
        \hline
        \end{tabular}
        \end{scriptsize}
    \end{minipage}
\end{table*}

\begin{table*}[t!]
    \centering
    \begin{scriptsize}
    \caption{Ablation of transformer block configuration.}
    \label{tab:trans_abl}
    \begin{tabular}{ccccccccc}
    \hline
        & 
        & 
        & \multicolumn{3}{c}{Ckt-Bench-101}
        & \multicolumn{3}{c}{Ckt-Bench-301} \\
        \cline{4-9}
          
        \multirow{-1.9}{*}{\makecell{Embedding \\ dimension}} 
        & \multirow{-1.9}{*}{\makecell{Number of \\ attention heads}}
        & \multirow{-1.9}{*}{\makecell{Number of \\transformer layers}}
        & Recon-Acc~(\%)
        & Validity~(\%)
        & Novelty~(\%)
        & Recon-Acc~(\%)
        & Validity~(\%)
        & Novelty~(\%)
        
        \\ \hline
          
        64
        & 4
        & 4
        & 96.80
        & 97.60
        & \textbf{95.91}
        & \underline{99.87}
        & 96.30
        & 90.68 \\
    
        64
        & 8
        & 4
        & \underline{97.30}
        & 97.70
        & 92.43
        & \textbf{99.99}
        & 97.70
        & 91.96 \\
    
        64
        & 8
        & 8
        & 78.40
        & \textbf{99.10}
        & 90.74
        & 99.76
        & \underline{98.40}
        & 90.88 \\
    
        128
        & 4
        & 4
        & 94.60
        & 96.80
        & 91.08
        & \textbf{99.99}
        & 96.70
        & 91.96 \\
    
        \rowcolor{aliceblue}
        128
        & 8
        & 4
        & \textbf{99.99}
        & \underline{98.70}
        & \underline{94.95}
        & \textbf{99.99}
        & \underline{98.40}
        & \underline{93.11} \\
    
        128
        & 8
        & 8
        & 78.30
        & 98.20
        & 91.28
        & 99.74
        & 98.30
        & 90.65 \\
    
        256
        & 4
        & 4
        & 86.60
        & 97.60
        & 93.47
        & \textbf{99.99}
        & 96.30
        & \textbf{94.61} \\
    
        256
        & 8
        & 4
        & 96.30
        & 97.70
        & 94.09
        & \textbf{99.99}
        & 97.20
        & 91.79 \\
    
        256
        & 8
        & 8
        & 81.20
        & 98.40
        & 91.28
        & 99.76
        & 97.80
        & 92.24 \\
    
        256
        & 16
        & 8
        & 78.10
        & 98.40
        & 93.31
        & 99.79
        & \textbf{98.90}
        & 91.83 \\
    
        256
        & 16
        & 16
        & 0.90
        & 94.50
        & 54.44
        & 26.31
        & 96.40
        & 39.88 \\
    
    \hline
    \end{tabular}
    \end{scriptsize}
\end{table*}

\subsection{Reconstruction and unconditional circuit generation}
\label{sec:recon_uncond}
\par Since our approach is developed based on the datasets proposed by CktGNN, we adopt the same evaluation settings reconstruction and unconditional circuit generation tasks. We encode test circuits to latent vectors and decode them to measure the reconstruction accuracy (Recon-Acc) metric. We then sample 1000 latent vectors from the Gaussian distribution and decode them into circuits to evaluate validity and novelty. Specifically, the CktGNN baseline results reported in this work correspond to the latest version of their implementation, which includes updated datasets and performance results.

\par We conducted reconstruction and unconditional generation experiments on Ckt-Bench-101 and Ckt-Bench-301, respectively. The results, shown in Table~\ref{exp_recon_uncond}~\cite{cktgnn,pace}, compare CktGen with the previous state-of-the-art~\cite{cktgnn}. CktGen outperformed the state-of-the-art in Recon-Acc and validity. In terms of the novelty metric, although the baselines achieved higher scores, they incorrectly classified some invalid circuits as novel ones. These results suggest that CktGen effectively captures the intricate patterns within circuit topologies, thereby enhancing the validity of the circuit generation process.

\par We also conducted an ablation study of $\lambda_{\text{type}}$ and $\lambda_{\text{pos}}$ (introduced in Section~\ref{sec:archi}) on Ckt-Bench-101 (Tables~\ref{tab:abl_recon_101}) and Ckt-Bench-301 (Tables~\ref{tab:abl_recon_301}). These experiments aimed to identify the optimal weights for $\mathcal{L}_{\text{type}}$ and $\mathcal{L}_{\text{pos}}$. For $\lambda_{\text{type}}$, we varied it from $0.1$ to $1$, and the interval is $0.1$. For $\lambda_{\text{pos}}$, we varied it from $0.01$ to $0.1$, and the interval is $0.01$. Additionally, we reported results without setting weights (\ie, $\lambda_{\text{type}}=\lambda_{\text{pos}}=1$). For $\lambda_{\text{size}}$, we followed the value in CktGNN, setting it to $0.01$. In Ckt-Bench-101, a too-low weights caused a significant decline in Recon-Acc, while in Ckt-Bench-301, it stabilized around the optimal value. This demonstrates that our transformer-based architecture improves stability as the data scale increases. In summary, a very low weight severely impacts the learning ability of the model, while a too-high value also degrades performance. We adjusted the loss weight hyperparameters to $\lambda_{\text{type}}=0.5$ and $\lambda_{\text{pos}}=0.05$ for Ckt-Bench-101, and $\lambda_{\text{type}}=0.7$ and $\lambda_{\text{pos}}=0.07$ for Ckt-Bench-301, respectively.

Table~\ref{tab:trans_abl} presents the ablation study of transformer block configurations on the Ckt-Bench-101 and Ckt-Bench-301 datasets. We systematically varied the embedding dimension, number of attention heads, and number of transformer layers to assess their impact on Recon-Acc, validity, and novelty. The results show that increasing the embedding dimension and number of layers beyond a certain point does not necessarily improve performance and can actually degrade both Recon-Acc and novelty. Notably, the configuration with an embedding dimension of 128, eight heads, and four layers achieves the best overall trade-off, yielding the highest Recon-Acc~(99.99\%) and strong validity and novelty across both datasets. In contrast, models with excessively large embedding dimensions or deeper architectures~(\eg, 16 layers) suffer from significant drops in both Recon-Acc and novelty, indicating potential overfitting or optimization difficulties. These findings highlight the importance of balanced model capacity and demonstrate that a moderate configuration is most effective for our circuit generation tasks.

\section{Discussion}
\par Our results show that specification-conditioned generative modeling can serve as a foundation for scalable analog circuit synthesis. CktGen learns structured relations between specification targets and circuit topologies, demonstrating the potential to augment existing analog design workflows. However, closing the remaining gap toward manufacturable and interactive automation requires advances in evaluation accuracy, generalization to diverse devices, and deep integration with electronic design automation (EDA) tools. Below, we discuss key directions toward this goal.

\subsection{Closed-loop surrogate feedback for self-improving circuit generation}

\par Although surrogate evaluation enables efficient generative exploration, current statistical models often fail to capture fine-grained circuit errors and layout-dependent parasitics. Neural simulators such as INSIGHT~\cite{insight} provide richer physical behaviors than scalar performance prediction. Reinforcement-based refinement frameworks including AutoCircuit-RL~\cite{autocircuitrl} further demonstrate adaptive improvement of topologies toward design-rule satisfaction. Meanwhile, log-driven diagnosis and auto-debug~\cite{llmlogdebug}, as well as structured circuit reasoning benchmarks such as CIRCUIT~\cite{circuitbenchmark}, encourage continuous model evolution under realistic verification. The recent SPICEPilot framework~\cite{spicepilot} also shows promise for coupling code generation with simulation feedback. Together, these advances point toward a closed-loop paradigm in which generation, evaluation, and constraint learning mutually reinforce one another. Within such loops, CktGen serves as a topology generator whose latent manifold can be progressively aligned to silicon-validated behaviors.

\subsection{Unified generative modeling across different analog circuit types}

\par Expanding to more diverse analog circuit types requires unified representations. Code-based and topology-based generation frameworks (\eg, AnalogCoder~\cite{analogcoder,analogcoderpro}, LaMAGIC~\cite{lamagic,lamagic2}) demonstrate that language priors can support diverse architectures. AnalogGenie~\cite{analoggenie} adopts a sequence-based topology generation paradigm that is conceptually aligned with ours, enabling the synthesis of diverse analog blocks such as low-dropout regulators~(LDOs), comparators, low-noise amplifiers~(LNAs), mixers, and voltage-controlled oscillators~(VCOs). Dataset and benchmark contributions~\cite{amsnetkg,autospice} improve robustness against distribution shift. Multi-agent and hierarchical modeling~\cite{analogxpert,ampagent} expand support to more complex analog systems, and foundation-model advances~\cite{analogseeker,adollm} highlight generalization across technology nodes. Yet, diverse specification semantics remain a key barrier to unified generation. CktGen offers a specification-grounded latent space capable of spanning multiple circuit types. We suggest encoding circuit type information (\eg, with one-hot vectors) and standardizing specification vectors via zero-padding. Strengthening its conditioning through compact type embedding and broader cross-domain datasets could enable a more universal generator with robust generalization across heterogeneous analog building blocks.

\subsection{Large language model assisted co-design within industrial EDA ecosystems}
\par For practical impact, integrating generative models into existing EDA workflows is essential for scalable deployment. Large language model~(LLM)-based assistants demonstrate complementary strengths in parameter design~\cite{llmadaisizing} and document-driven intent extraction~\cite{doceda}, and increasingly support standard workflows in SPICE-oriented generation~\cite{autospice}, netlist editing~\cite{schemato}, and layout verification~\cite{glayout}. Layout-centric copilots~\cite{glayout,layoutcopilot} and constraint-aware editing frameworks~\cite{interactiveanaloglayout,intlayoutedit} further indicate the potential to propagate symmetry and matching intent into physical implementation. Moreover, incorporating knowledge from design rule checking (DRC), layout versus schematic (LVS), and routing logs~\cite{llmenaobo,edaawarertl} allows the continuous refinement of design heuristics tailored to specific process design kits~(PDKs) and technology nodes. Taken together, these trajectories support a human-in-the-loop co-design paradigm in which LLM agents interact directly with commercial EDA environments. In this setting, CktGen serves as the generative backbone, supplying high-diversity, specification-consistent candidates that LLM copilots can adapt through native tool feedback toward manufacturable, low-effort implementation. Advanced knowledge representation techniques could further enable more sophisticated reasoning and knowledge integration within these LLM-assisted EDA workflows~\cite{knowledge_representation}.

\section{Conclusions}
\par We have demonstrated that specification-conditioned generative modeling effectively maps performance requirements to diverse and valid analog circuit implementations. Through joint representation learning, CktGen learns robust one-to-many mappings that enable flexible adaptation to changing design requirements without retraining, achieving superior performance in specification-conditioned generation compared with existing methods. MAB-based test-time optimization enables efficient search for high-performance circuits that meet target specifications while obtaining higher FoM, demonstrating significant improvements in automated design tasks. CktGen also achieves high Recon-Acc and validity in unconditional generation tasks. While the current approach demonstrates promising results, future work should address limitations, including fine-grained circuit behaviors, generalization to broader circuit types, and integration with industrial EDA workflows. By reformulating analog circuit synthesis as a conditional generation problem, this work establishes a new paradigm that overcomes the flexibility limitations of traditional optimization-based approaches, positioning generative artificial intelligence as a transformative tool for accelerating analog design automation.

\section*{Acknowledgements}
This work was supported by the National Natural Science Foundation of China (62472381 and U2336212), the Fundamental Research Funds for the Central Universities (226-2025-00080), and the Fundamental Research Funds for the Zhejiang Provincial Universities (226-2024-00208).

\section*{Data availability}
The source codes and model weights trained for specification-conditioned circuit generation, automated design, reconstruction, and unconditional generation, which were used to produce the results in this paper are available at \href{https://github.com/hhyxx/CktGen}{https://github.com/yuxuan-hou-x/CktGen}. Due to file size limitations, the weights are hosted on external cloud storage, with download links provided in the repository.


\bibliographystyle{unsrt}
\bibliography{cktgen}

@article{knowledge_representation,
  title={Multiple knowledge representation for big data artificial intelligence: framework, applications, and case studies},
  author={Yang, Yi and Zhuang, Yueting and Pan, Yunhe},
  journal={Frontiers of Information Technology \& Electronic Engineering},
  volume={22},
  number={12},
  pages={1551--1558},
  year={2021},
  publisher={Springer}
}

@article{analogxpert,
  title={AnalogXpert: Automating Analog Topology Synthesis by Incorporating Circuit Design Expertise into Large Language Models},
  author={Zhang, Haoyi and Sun, Shizhao and Lin, Yibo and Wang, Runsheng and Bian, Jiang},
  journal={arXiv preprint arXiv:2412.19824},
  year={2024}
}

@inproceedings{spicepilot,
  title={Spicepilot: Navigating spice code generation and simulation with ai guidance},
  author={Vungarala, Deepak and Alam, Sakila and Ghosh, Arnob and Angizi, Shaahin},
  booktitle={2024 IEEE International Conference on Rebooting Computing (ICRC)},
  pages={1--6},
  year={2024},
  organization={IEEE}
}

@article{schemato,
  title={Schemato – An LLM for Netlist-to-Schematic Conversion},
  author={Matsuo, Ryoga and Uhlich, Stefan and Venkitaraman, Arun and Bonetti, Andrea and Hsieh, Chia-Yu and Momeni, Ali and Mauch, Lukas and Capone, Augusto and Ohbuchi, Eisaku and Servadei, Lorenzo},
  journal={arXiv preprint arXiv:2411.13899},
  year={2024}
}

@article{ampagent,
  title={Ampagent: An llm-based multi-agent system for multi-stage amplifier schematic design from literature for process and performance porting},
  author={Liu, Chengjie and Chen, Weiyu and Peng, Anlan and Du, Yuan and Du, Li and Yang, Jun},
  journal={arXiv preprint arXiv:2409.14739},
  year={2024}
}

@inproceedings{analogcoder,
  title={Analogcoder: Analog circuit design via training-free code generation},
  author={Lai, Yao and Lee, Sungyoung and Chen, Guojin and Poddar, Souradip and Hu, Mengkang and Pan, David Z and Luo, Ping},
  booktitle={Proceedings of the AAAI Conference on Artificial Intelligence},
  volume={39},
  number={1},
  pages={379--387},
  year={2025}
}

@article{analogcoderpro,
  title={AnalogCoder-Pro: Unifying Analog Circuit Generation and Optimization via Multi-modal LLMs},
  author={Lai, Yao and Poddar, Souradip and Lee, Sungyoung and Chen, Guojin and Hu, Mengkang and Yu, Bei and Luo, Ping and Pan, David Z},
  journal={arXiv preprint arXiv:2508.02518},
  year={2025}
}

@inproceedings{lamagic,
  title={LaMAGIC: Language-Model-based Topology Generation for Analog Integrated Circuits},
  author={Chang, Chen-Chia and Shen, Yikang and Fan, Shaoze and Li, Jing and Zhang, Shun and Cao, Ningyuan and Chen, Yiran and Zhang, Xin},
  booktitle={International Conference on Machine Learning},
  pages={6253--6262},
  year={2024},
  organization={PMLR}
}

@inproceedings{lamagic2,
  title={LaMAGIC2: Advanced Circuit Formulations for Language Model-Based Analog Topology Generation},
  author={Chang, Chen-Chia and Lin, Wan-Hsuan and Shen, Yikang and Chen, Yiran and Zhang, Xin},
  booktitle={Forty-second International Conference on Machine Learning}
}

@article{llmlogdebug,
  title={LLM-Powered EDA Log Analysis for Effective Design Debugging},
  author={Kanagal, Rohit},
  year={2025}
}

@inproceedings{edaawarertl,
  title={EDA-Aware RTL Generation with Large Language Models},
  author={ul Islam, Mubashir and Sami, Humza and Gaillardon, Pierre-Emmanuel and Tenace, Valerio},
  booktitle={2025 Design, Automation \& Test in Europe Conference (DATE)},
  pages={1--6},
  year={2025},
  organization={IEEE}
}

@article{autospice,
  title={Auto-spice: Leveraging llms for dataset creation via automated spice netlist extraction from analog circuit diagrams},
  author={Bhandari, Jitendra and Bhat, Vineet and He, Yuheng and Garg, Siddharth and Rahmani, Hamed and Karri, Ramesh},
  journal={arXiv e-prints},
  pages={arXiv--2411},
  year={2024}
}

@article{llmenaobo,
  title={LLM-enhanced Bayesian optimization for efficient analog layout constraint generation},
  author={Chen, Guojin and Zhu, Keren and Kim, Seunggeun and Zhu, Hanqing and Lai, Yao and Yu, Bei and Pan, David Z},
  journal={arXiv preprint arXiv:2406.05250},
  year={2024}
}

@article{layoutcopilot,
  title={Layoutcopilot: An llm-powered multi-agent collaborative framework for interactive analog layout design},
  author={Liu, Bingyang and Zhang, Haoyi and Gao, Xiaohan and Kong, Zichen and Tang, Xiyuan and Lin, Yibo and Wang, Runsheng and Huang, Ru},
  journal={IEEE Transactions on Computer-Aided Design of Integrated Circuits and Systems},
  year={2025},
  publisher={IEEE}
}

@article{llmadaisizing,
  title={LLM-based AI Agent for Sizing of Analog and Mixed Signal Circuit},
  author={Liu, Chang and Olowe, Emmanuel A and Chitnis, Danial},
  journal={arXiv preprint arXiv:2504.11497},
  year={2025}
}

@inproceedings{adollm,
  title={Ado-llm: Analog design bayesian optimization with in-context learning of large language models},
  author={Yin, Yuxuan and Wang, Yu and Xu, Boxun and Li, Peng},
  booktitle={Proceedings of the 43rd IEEE/ACM International Conference on Computer-Aided Design},
  pages={1--9},
  year={2024}
}

@article{analogseeker,
  title={AnalogSeeker: An Open-source Foundation Language Model for Analog Circuit Design},
  author={Chen, Zihao and Zhuang, Ji and Shen, Jinyi and Ke, Xiaoyue and Yang, Xinyi and Zhou, Mingjie and Du, Zhuoyao and Yan, Xu and Wu, Zhouyang and Xu, Zhenyu and others},
  journal={arXiv preprint arXiv:2508.10409},
  year={2025}
}

@inproceedings{glayout,
  title={Human language to analog layout using Glayout layout automation framework},
  author={Hammoud, Ali and Goyal, Chetanya and Pathen, Sakib and Dai, Arlene and Li, Anhang and Kielian, Gregory and Saligane, Mehdi},
  booktitle={Proceedings of the 2024 ACM/IEEE International Symposium on Machine Learning for CAD},
  pages={1--7},
  year={2024}
}

@inproceedings{autocircuitrl,
  title={AUTOCIRCUIT-RL: Reinforcement Learning-Driven LLM for Automated Circuit Topology Generation},
  author={Vijayaraghavan, Prashanth and Shi, Luyao and Degan, Ehsan and Mukherjee, Vandana and Zhang, Xin},
  booktitle={Forty-second International Conference on Machine Learning}
}

@article{amsnetkg,
  title={AMSnet-KG: A netlist dataset for LLM-based AMS circuit auto-design using knowledge graph RAG},
  author={Shi, Yichen and Tao, Zhuofu and Gao, Yuhao and Zhou, Tianjia and Chang, Cheng and Wang, Yaxin and Chen, Bingyu and Zhang, Genhao and Liu, Alvin and Yu, Zhiping and others},
  journal={ACM Transactions on Design Automation of Electronic Systems},
  volume={30},
  number={6},
  pages={1--37},
  year={2025},
  publisher={ACM New York, NY}
}

@article{circuitbenchmark,
  title={CIRCUIT: A Benchmark for Circuit Interpretation and Reasoning Capabilities of LLMs},
  author={Skelic, Lejla and Xu, Yan and Cox, Matthew and Lu, Wenjie and Yu, Tao and Han, Ruonan},
  journal={arXiv preprint arXiv:2502.07980},
  year={2025}
}

@article{intlayoutedit,
  title={Interactive analog layout editing with instant placement and routing legalization},
  author={Gao, Xiaohan and Zhang, Haoyi and Liu, Mingjie and Shen, Linxiao and Pan, David Z and Lin, Yibo and Wang, Runsheng and Huang, Ru},
  journal={IEEE Transactions on Computer-Aided Design of Integrated Circuits and Systems},
  volume={42},
  number={3},
  pages={698--711},
  year={2022},
  publisher={IEEE}
}

@inproceedings{interactiveanaloglayout,
  title={Intelligent and interactive analog layout design automation},
  author={Lin, Yibo and Gao, Xiaohan and Zhang, Haoyi and Wang, Runsheng and Huang, Ru},
  booktitle={2022 IEEE 16th International Conference on Solid-State \& Integrated Circuit Technology (ICSICT)},
  pages={1--4},
  year={2022},
  organization={IEEE}
}

@article{insight,
  title={Insight: Universal neural simulator for analog circuits harnessing autoregressive transformers},
  author={Poddar, Souradip and Oh, Youngmin and Lai, Yao and Zhu, Hanqing and Hwang, Bosun and Pan, David Z},
  journal={arXiv preprint arXiv:2407.07346},
  year={2024}
}

@article{doceda,
  title={DocEDA: Automated extraction and design of analog circuits from documents with large language model},
  author={Chen, Hong Cai and Wu, Longchang and Gao, Ming and Shen, Lingrui and Zhong, Jiarui and Xu, Yipin},
  journal={arXiv preprint arXiv:2412.05301},
  year={2024}
}

@inproceedings{ldt,
  title={Scalable diffusion models with transformers},
  author={Peebles, William and Xie, Saining},
  booktitle={Proceedings of the IEEE/CVF international conference on computer vision},
  pages={4195--4205},
  year={2023}
}

@inproceedings{cvaegan,
  title={CVAE-GAN: fine-grained image generation through asymmetric training},
  author={Bao, Jianmin and Chen, Dong and Wen, Fang and Li, Houqiang and Hua, Gang},
  booktitle={Proceedings of the IEEE international conference on computer vision},
  pages={2745--2754},
  year={2017}
}

@article{review_conv,
  title={An incipient fault diagnosis method based on complex convolutional self-attention autoencoder for analog circuits},
  author={Gao, Tianyu and Yang, Jingli and Jiang, Shouda and Li, Ye},
  journal={IEEE Transactions on Industrial Electronics},
  volume={71},
  number={8},
  pages={9727--9736},
  year={2023},
  publisher={IEEE}
}

@article{review_vq,
  title={A novel fault detection model based on vector quantization sparse autoencoder for nonlinear complex systems},
  author={Gao, Tianyu and Yang, Jingli and Jiang, Shouda},
  journal={IEEE Transactions on Industrial Informatics},
  volume={19},
  number={3},
  pages={2693--2704},
  year={2022},
  publisher={IEEE}
}

@article{gin,
  title={How powerful are graph neural networks?},
  author={Xu, Keyulu and Hu, Weihua and Leskovec, Jure and Jegelka, Stefanie},
  journal={arXiv preprint arXiv:1810.00826},
  year={2018}
}

@article{analoggenie,
  title={AnalogGenie: A generative engine for automatic discovery of analog circuit topologies},
  author={Gao, Jian and Cao, Weidong and Yang, Junyi and Zhang, Xuan},
  journal={arXiv preprint arXiv:2503.00205},
  year={2025}
}

@article{simclr_v2,
  title={Big self-supervised models are strong semi-supervised learners},
  author={Chen, Ting and Kornblith, Simon and Swersky, Kevin and Norouzi, Mohammad and Hinton, Geoffrey E},
  journal={Advances in neural information processing systems},
  volume={33},
  pages={22243--22255},
  year={2020}
}

@article{moco_v2,
  title={Improved baselines with momentum contrastive learning},
  author={Chen, Xinlei and Fan, Haoqi and Girshick, Ross and He, Kaiming},
  journal={arXiv preprint arXiv:2003.04297},
  year={2020}
}

@article{zhang2022region,
  title={Region-level contrastive and consistency learning for semi-supervised semantic segmentation},
  author={Zhang, Jianrong and Wu, Tianyi and Ding, Chuanghao and Zhao, Hongwei and Guo, Guodong},
  journal={arXiv preprint arXiv:2204.13314},
  year={2022}
}

@inproceedings{t2m_gpt,
  title={Generating human motion from textual descriptions with discrete representations},
  author={Zhang, Jianrong and Zhang, Yangsong and Cun, Xiaodong and Zhang, Yong and Zhao, Hongwei and Lu, Hongtao and Shen, Xi and Shan, Ying},
  booktitle={Proceedings of the IEEE/CVF conference on computer vision and pattern recognition},
  pages={14730--14740},
  year={2023}
}

@article{surrogate,
  title={Surrogate-based analysis and optimization},
  author={Queipo, Nestor V and Haftka, Raphael T and Shyy, Wei and Goel, Tushar and Vaidyanathan, Rajkumar and Tucker, P Kevin},
  journal={Progress in aerospace sciences},
  volume={41},
  number={1},
  pages={1--28},
  year={2005},
  publisher={Elsevier}
}

@article{gnn,
  title={Semi-supervised classification with graph convolutional networks},
  author={Kipf, Thomas N and Welling, Max},
  journal={arXiv preprint arXiv:1609.02907},
  year={2016}
}

@article{adamw,
  title={Decoupled weight decay regularization},
  author={Loshchilov, I},
  journal={arXiv preprint arXiv:1711.05101},
  year={2017}
}

@article{spice,
  title={DELIGHT. SPICE: An optimization-based system for the design of integrated circuits},
  author={Nye, William and Riley, David C and Sangiovanni-Vincentelli, Alberto and Tits, Andre L},
  journal={IEEE Transactions on Computer-Aided Design of Integrated Circuits and Systems},
  volume={7},
  number={4},
  pages={501--519},
  year={1988},
  publisher={IEEE}
}

@inproceedings{temos,
  title={TEMOS: Generating diverse human motions from textual descriptions},
  author={Petrovich, Mathis and Black, Michael J and Varol, G{\"u}l},
  booktitle={European Conference on Computer Vision},
  pages={480--497},
  year={2022},
  organization={Springer}
}

@article{design_flow_survey,
  title={Computer-aided design of analog and mixed-signal integrated circuits},
  author={Gielen, Georges GE and Rutenbar, Rob A},
  journal={Proceedings of the IEEE},
  volume={88},
  number={12},
  pages={1825--1854},
  year={2000}, 
  publisher={IEEE}
}

@article{survey1,
  title={Machine learning for electronic design automation: A survey},
  author={Huang, Guyue and Hu, Jingbo and He, Yifan and Liu, Jialong and Ma, Mingyuan and Shen, Zhaoyang and Wu, Juejian and Xu, Yuanfan and Zhang, Hengrui and Zhong, Kai and others},
  journal={ACM Transactions on Design Automation of Electronic Systems (TODAES)},
  volume={26},
  number={5},
  pages={1--46},
  year={2021},
  publisher={ACM New York, NY}
}

@inproceedings{survey2,
  title={ML for Analog Design: Good Progress, But More to Do},
  author={Nikoli{\'c}, Borivoje},
  booktitle={Proceedings of the 2022 ACM/IEEE Workshop on Machine Learning for CAD},
  pages={53--54},
  year={2022}
}

@article{survey3,
  title={Automated topology synthesis of analog and RF integrated circuits: A survey},
  author={Sorkhabi, Samin Ebrahim and Zhang, Lihong},
  journal={Integration},
  volume={56},
  pages={128--138},
  year={2017},
  publisher={Elsevier}
}

@article{vae,
  title={Auto-encoding variational bayes},
  author={Kingma, Diederik P and Welling, Max},
  journal={arXiv preprint arXiv:1312.6114},
  year={2013}
}

@article{transformer,
  title={Attention is all you need},
  author={Vaswani, Ashish and Shazeer, Noam and Parmar, Niki and Uszkoreit, Jakob and Jones, Llion and Gomez, Aidan N and Kaiser, {\L}ukasz and Polosukhin, Illia},
  journal={Advances in neural information processing systems},
  volume={30},
  year={2017}
}

@article{infonce,
  title={Representation learning with contrastive predictive coding},
  author={Oord, Aaron van den and Li, Yazhe and Vinyals, Oriol},
  journal={arXiv preprint arXiv:1807.03748},
  year={2018}
}

@inproceedings{cktgnn,
    title={Ckt{GNN}:  Circuit Graph Neural Network for Electronic Design Automation},
    author={Zehao Dong and Weidong Cao and Muhan Zhang and Dacheng Tao and Yixin Chen and Xuan Zhang},
    booktitle={The Eleventh International Conference on Learning Representations },
    year={2023},
    url={https://openreview.net/forum?id=NE2911Kq1sp}
}

@inproceedings{pace,
  title={Pace: A parallelizable computation encoder for directed acyclic graphs},
  author={Dong, Zehao and Zhang, Muhan and Li, Fuhai and Chen, Yixin},
  booktitle={International Conference on Machine Learning},
  pages={5360--5377},
  year={2022},
  organization={PMLR}
}

@INPROCEEDINGS{topo_rule_1,
  author={Chavez, J. and Torralba, A. and Franquelo, L.G.},
  booktitle={Proceedings of IEEE International Symposium on Circuits and Systems - ISCAS '94}, 
  title={A fuzzy-logic based tool for topology selection in analog synthesis}, 
  year={1994},
  volume={1},
  number={},
  pages={367-370 vol.1},
  doi={10.1109/ISCAS.1994.408873}}

@article{topo_rule_2,
  title={FASY: A fuzzy-logic based tool for analog synthesis},
  author={Torralba, Antonio and Chavez, Jorge and Franquelo, Leopoldo Garc{\'\i}a},
  journal={IEEE Transactions on Computer-Aided Design of Integrated Circuits and Systems},
  volume={15},
  number={7},
  pages={705--715},
  year={1996},
  publisher={IEEE}
}

@inproceedings{topo_rule_3,
  title={A graph grammar based approach to automated multi-objective analog circuit design},
  author={Das, Angan and Vemuri, Ranga},
  booktitle={2009 Design, Automation \& Test in Europe Conference \& Exhibition},
  pages={700--705},
  year={2009},
  organization={IEEE}
}

@inproceedings{topo_rule_4,
  title={Graph-grammar-based analog circuit topology synthesis},
  author={Zhao, Zhenxin and Zhang, Lihong},
  booktitle={2019 IEEE International Symposium on Circuits and Systems (ISCAS)},
  pages={1--5},
  year={2019},
  organization={IEEE}
}

@article{topo_heu_2,
  title={A synthesis system for analog circuits based on evolutionary search and topological reuse},
  author={Dastidar, Tathagato Rai and Chakrabarti, PP and Ray, Partha},
  journal={IEEE Transactions on evolutionary computation},
  volume={9},
  number={2},
  pages={211--224},
  year={2005},
  publisher={IEEE}
}

@article{topo_heu_3,
  title={Analog genetic encoding for the evolution of circuits and networks},
  author={Mattiussi, Claudio and Floreano, Dario},
  journal={IEEE Transactions on evolutionary computation},
  volume={11},
  number={5},
  pages={596--607},
  year={2007},
  publisher={IEEE}
}

@article{topo_heu_5,
  title={Trustworthy genetic programming-based synthesis of analog circuit topologies using hierarchical domain-specific building blocks},
  author={McConaghy, Trent and Palmers, Pieter and Steyaert, Michiel and Gielen, Georges GE},
  journal={IEEE Transactions on Evolutionary Computation},
  volume={15},
  number={4},
  pages={557--570},
  year={2011},
  publisher={IEEE}
}

@article{topo_heu_7,
  title={Analog circuit topology synthesis by means of evolutionary computation},
  author={Rojec, {\v{Z}}iga and B{\H{u}}rmen, {\'A}rp{\'a}d and Fajfar, Iztok},
  journal={Engineering Applications of Artificial Intelligence},
  volume={80},
  pages={48--65},
  year={2019},
  publisher={Elsevier}
}

@inproceedings{topo_bo,
  title={Topology optimization of operational amplifier in continuous space via graph embedding},
  author={Lu, Jialin and Lei, Liangbo and Yang, Fan and Shang, Li and Zeng, Xuan},
  booktitle={2022 Design, Automation \& Test in Europe Conference \& Exhibition (DATE)},
  pages={142--147},
  year={2022},
  organization={IEEE}
}

@inproceedings{topo_rl_1,
  title={Automatic analog schematic diagram generation based on building block classification and reinforcement learning},
  author={Hsu, Hung-Yun and Lin, Mark Po-Hung},
  booktitle={Proceedings of the 2022 ACM/IEEE Workshop on Machine Learning for CAD},
  pages={43--48},
  year={2022}
}

@article{topo_rl_2,
  title={Analog integrated circuit topology synthesis with deep reinforcement learning},
  author={Zhao, Zhenxin and Zhang, Lihong},
  journal={IEEE Transactions on Computer-Aided Design of Integrated Circuits and Systems},
  volume={41},
  number={12},
  pages={5138--5151},
  year={2022},
  publisher={IEEE}
}

@inproceedings{topo_rl_3,
  title={TOTAL: Topology Optimization of Operational Amplifier via Reinforcement Learning},
  author={Chen, Zihao and Meng, Songlei and Yang, Fan and Shang, Li and Zeng, Xuan},
  booktitle={2023 24th International Symposium on Quality Electronic Design (ISQED)},
  pages={1--8},
  year={2023},
  organization={IEEE}
}

@article{sz_know_1,
  title={IDAC: An interactive design tool for analog CMOS circuits},
  author={Degrauwe, Marc GR and Nys, Olivier and Dijkstra, Evert and Rijmenants, Jef and Bitz, Serge and Goffart, Bernard LAG and Vittoz, Eric A and Cserveny, Stefan and Meixenberger, Christian and Van Der Stappen, G and others},
  journal={IEEE Journal of solid-state circuits},
  volume={22},
  number={6},
  pages={1106--1116},
  year={1987},
  publisher={IEEE}
}

@article{sz_know_2,
  title={OASYS: A framework for analog circuit synthesis},
  author={Harjani, Ramesh and Rutenbar, Rob A and Carley, L Richard},
  journal={IEEE Transactions on Computer-Aided Design of Integrated Circuits and Systems},
  volume={8},
  number={12},
  pages={1247--1266},
  year={1989},
  publisher={IEEE}
}

@article{sz_know_3,
  title={Analog IC design automation. II. Automated circuit correction by qualitative reasoning},
  author={Makris, Costas A and Toumazou, Christofer},
  journal={IEEE transactions on computer-aided design of integrated circuits and systems},
  volume={14},
  number={2},
  pages={239--254},
  year={1995},
  publisher={IEEE}
}

@article{sz_heu_2,
  title={Analog circuit design optimization based on symbolic simulation and simulated annealing},
  author={Gielen, Georges GE and Walscharts, Herman CC and Sansen, Willy MC},
  journal={IEEE Journal of solid-state circuits},
  volume={25},
  number={3},
  pages={707--713},
  year={1990},
  publisher={IEEE}
}

@inproceedings{sz_heu_3,
  title={Swarm intelligence based sizing methodology for CMOS operational amplifier},
  author={Vural, Revna Acar and Yildirim, Tulay},
  booktitle={2011 IEEE 12th International Symposium on Computational Intelligence and Informatics (CINTI)},
  pages={525--528},
  year={2011},
  organization={IEEE}
}

@inproceedings{sz_bo_1,
  title={Batch Bayesian optimization via multi-objective acquisition ensemble for automated analog circuit design},
  author={Lyu, Wenlong and Yang, Fan and Yan, Changhao and Zhou, Dian and Zeng, Xuan},
  booktitle={International conference on machine learning},
  pages={3306--3314},
  year={2018},
  organization={PMLR}
}

@inproceedings{sz_bo_2,
  title={Bayesian optimization approach for analog circuit synthesis using neural network},
  author={Zhang, Shuhan and Lyu, Wenlong and Yang, Fan and Yan, Changhao and Zhou, Dian and Zeng, Xuan},
  booktitle={2019 Design, Automation \& Test in Europe Conference \& Exhibition (DATE)},
  pages={1463--1468},
  year={2019},
  organization={IEEE}
}

@inproceedings{sz_bo_3,
  title={Multi-objective bayesian optimization for analog/rf circuit synthesis},
  author={Lyu, Wenlong and Yang, Fan and Yan, Changhao and Zhou, Dian and Zeng, Xuan},
  booktitle={Proceedings of the 55th Annual Design Automation Conference},
  pages={1--6},
  year={2018}
}

@article{sz_nn_1,
  title={An artificial neural network assisted optimization system for analog design space exploration},
  author={Li, Yaping and Wang, Yong and Li, Yusong and Zhou, Ranran and Lin, Zhaojun},
  journal={IEEE Transactions on Computer-Aided Design of Integrated Circuits and Systems},
  volume={39},
  number={10},
  pages={2640--2653},
  year={2019},
  publisher={IEEE}
}

@article{sz_nn_2,
  author={Budak, Ahmet Faruk and Gandara, Miguel and Shi, Wei and Pan, David Z. and Sun, Nan and Liu, Bo},
  journal={IEEE Transactions on Computer-Aided Design of Integrated Circuits and Systems}, 
  title={An Efficient Analog Circuit Sizing Method Based on Machine Learning Assisted Global Optimization}, 
  year={2022},
  volume={41},
  number={5},
  pages={1209-1221},
  doi={10.1109/TCAD.2021.3081405}
}

@inproceedings{sz_rl_1,
  title={GCN-RL circuit designer: Transferable transistor sizing with graph neural networks and reinforcement learning},
  author={Wang, Hanrui and Wang, Kuan and Yang, Jiacheng and Shen, Linxiao and Sun, Nan and Lee, Hae-Seung and Han, Song},
  booktitle={2020 57th ACM/IEEE Design Automation Conference (DAC)},
  pages={1--6},
  year={2020},
  organization={IEEE}
}

@inproceedings{sz_rl_2,
  title={Domain knowledge-based automated analog circuit design with deep reinforcement},
  author={Cao, Weidong and Benosman, Mouhacine and Zhang, Xuan and Ma, Rui},
  booktitle={Proceedings of the 59th ACM/IEEE Design Automation Conference},
  pages={1015--1020},
  year={2022}
}

@inproceedings{sz_rl_3,
  title={RoSE: Robust Analog Circuit Parameter Optimization with Sampling-Efficient Reinforcement Learning},
  author={Gao, Jian and Cao, Weidong and Zhang, Xuan},
  booktitle={2023 60th ACM/IEEE Design Automation Conference (DAC)},
  pages={1--6},
  year={2023},
  organization={IEEE}
}

@article{dvae,
  title={D-vae: A variational autoencoder for directed acyclic graphs},
  author={Zhang, Muhan and Jiang, Shali and Cui, Zhicheng and Garnett, Roman and Chen, Yixin},
  journal={Advances in neural information processing systems},
  volume={32},
  year={2019}
}

@article{syn_heu_1,
  title={Automated synthesis of analog electrical circuits by means of genetic programming},
  author={Koza, John R. and Bennett, Forrest H and Andre, David and Keane, Martin A. and Dunlap, Frank},
  journal={IEEE Transactions on evolutionary computation},
  volume={1},
  number={2},
  pages={109--128},
  year={1997},
  publisher={IEEE}
}

@article{syn_heu_2,
  title={FEATS: Framework for explorative analog topology synthesis},
  author={Meissner, Markus and Hedrich, Lars},
  journal={IEEE Transactions on Computer-Aided Design of Integrated Circuits and Systems},
  volume={34},
  number={2},
  pages={213--226},
  year={2014},
  publisher={IEEE}
}

@inproceedings{wicked,
  title={WiCkeD: Analog circuit synthesis incorporating mismatch},
  author={Antreich, Kurt and Eckmueller, Josef and Graeb, Helmut and Pronath, Michael and Schenkel, Frank and Schwencker, Robert and Zizala, Stephan},
  booktitle={Proceedings of the IEEE 2000 Custom Integrated Circuits Conference (Cat. No. 00CH37044)},
  pages={511--514},
  year={2000},
  organization={IEEE}
}

@article{syn_lrn_1,
  title={Angel: Fully-automated analog circuit generator using a neural network assisted semi-supervised learning approach},
  author={Fayazi, Morteza and Taba, Morteza Tavakoli and Afshari, Ehsan and Dreslinski, Ronald},
  journal={IEEE Transactions on Circuits and Systems I: Regular Papers},
  year={2023},
  publisher={IEEE}
}

@article{syn_lrn_2,
  title={Automatic Op-Amp Generation From Specification to Layout},
  author={Lu, Jialin and Lei, Liangbo and Huang, Jiangli and Yang, Fan and Shang, Li and Zeng, Xuan},
  journal={IEEE Transactions on Computer-Aided Design of Integrated Circuits and Systems},
  year={2023},
  publisher={IEEE}
}

@article{magical,
  title={MAGICAL: An open-source fully automated analog IC layout system from netlist to GDSII},
  author={Chen, Hao and Liu, Mingjie and Xu, Biying and Zhu, Keren and Tang, Xiyuan and Li, Shaolan and Lin, Yibo and Sun, Nan and Pan, David Z},
  journal={IEEE Design \& Test},
  volume={38},
  number={2},
  pages={19--26},
  year={2020},
  publisher={IEEE}
}

@article{gm/id,
  title={A g/sub m//I/sub D/based methodology for the design of CMOS analog circuits and its application to the synthesis of a silicon-on-insulator micropower OTA},
  author={Silveira, Fernando and Flandre, Denis and Jespers, Paul GA},
  journal={IEEE journal of solid-state circuits},
  volume={31},
  number={9},
  pages={1314--1319},
  year={1996},
  publisher={IEEE}
}

@inproceedings{syn_lrn_3,
  title={MACRO: Multi-agent Reinforcement Learning-based Cross-layer Optimization of Operational Amplifier},
  author={Chen, Zihao and Meng, Songlei and Yang, Fan and Shang, Li and Zeng, Xuan},
  booktitle={2024 29th Asia and South Pacific Design Automation Conference (ASP-DAC)},
  pages={423--428},
  year={2024},
  organization={IEEE}
}

@article{fid,
  title={Gans trained by a two time-scale update rule converge to a local nash equilibrium},
  author={Heusel, Martin and Ramsauer, Hubert and Unterthiner, Thomas and Nessler, Bernhard and Hochreiter, Sepp},
  journal={Advances in neural information processing systems},
  volume={30},
  year={2017}
}

\newpage
\appendix
\onecolumn
\appendixpage

\begin{spacing}{1}
\section*{Contents}

\startcontents[appendices]
\printcontents[appendices]{}{-1}{\setcounter{tocdepth}{2}}
\end{spacing}

\section{Statistical analysis of the joint specification class distributions}
\label{appd:sec:a}
\par To analyze the statistical properties of joint specification class distributions in the Open-Circuit-Benchmark datasets, we conduct an analysis of the preprocessed specification space. We use a mapping function that projects each multi-dimensional specification vector into a scalar representation $y$:
\begin{small}%
\begin{equation}
    y = s_{\text{Gain}} + 10^{3} \times s_{\text{BW}} + 10^{6} \times s_{\text{PM}}
\end{equation}
\end{small}%
where the coefficients are designed to ensure distinct contributions from each dimension, thereby preventing collisions between specification combinations.

\par
\begin{figure*}[b!]
  \centering
  \centerline{\includegraphics[width=\textwidth]{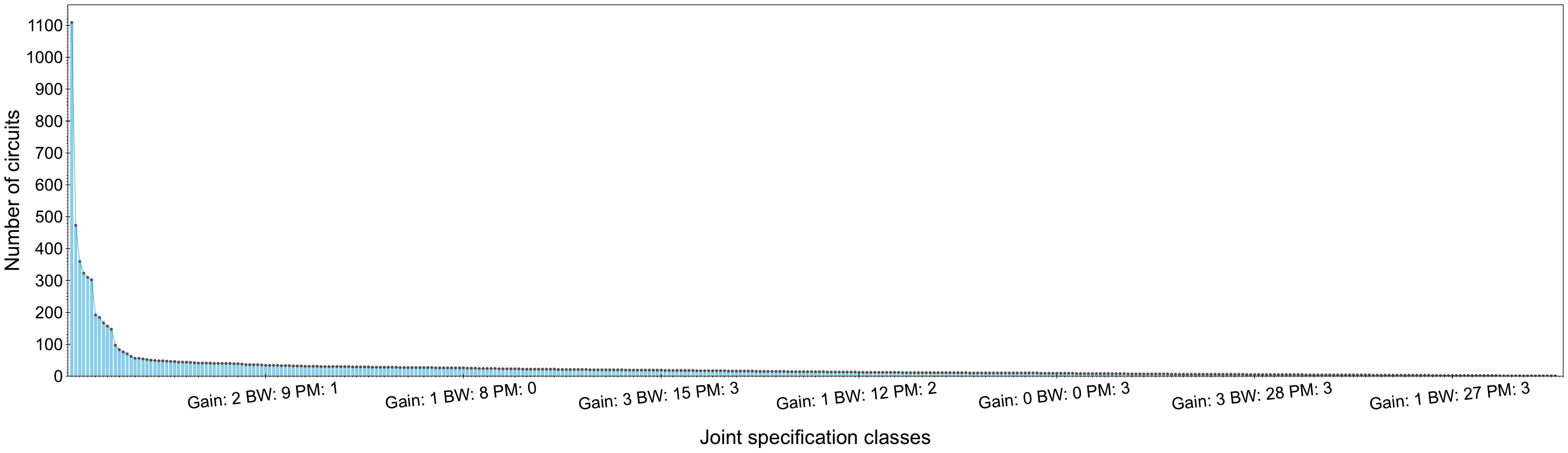}}
  \caption{Distribution of joint specification classes in Ckt-Bench-101.}
  \label{fig:101_dist}
\end{figure*}

\begin{figure*}[h!]
  \centering
  \centerline{\includegraphics[width=\textwidth]{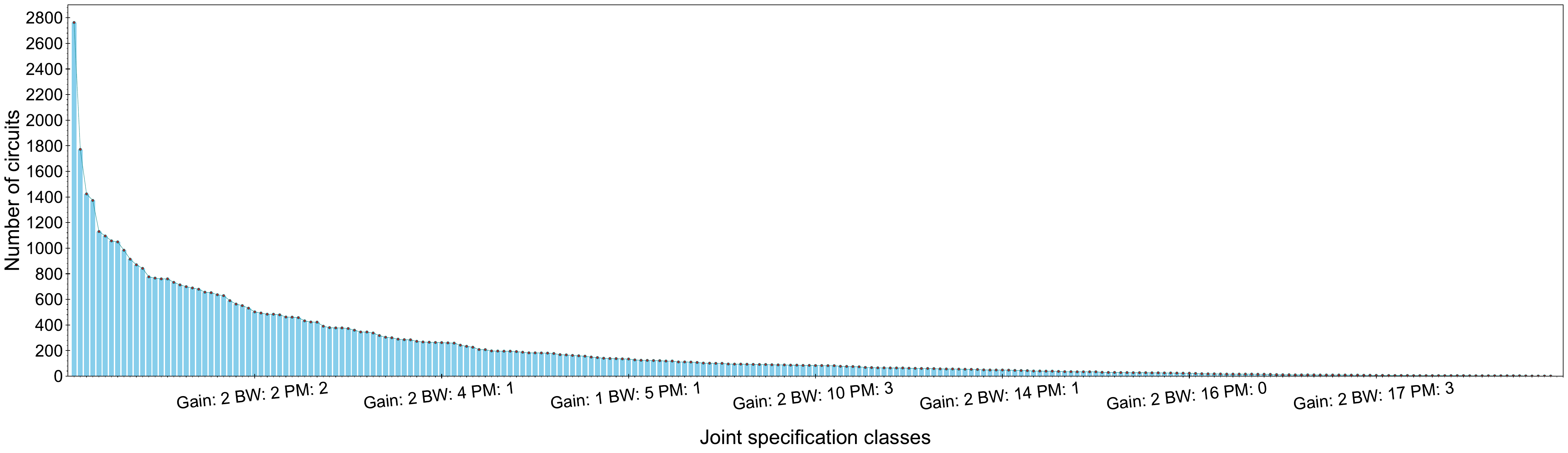}}
  \caption{Distribution of joint specification classes in Ckt-Bench-301.}
  \label{fig:301_dist}
\end{figure*}

 We compute the frequency distribution by counting circuit instances sharing identical $y$ values. Joint specification classes are sorted in descending order of frequency. Fig.~\ref{fig:101_dist} and Fig.~\ref{fig:301_dist} illustrate the distributions for Ckt-Bench-101 and Ckt-Bench-301, respectively. The horizontal axis represents the number of circuit instances per joint specification class, while the vertical axis denotes distinct joint specification classes indexed by $y$. 

\par The distributions reveal significant class imbalance, with some joint specification classes containing hundreds of implementations while others have only a few. Importantly, for any given joint specification class, multiple valid circuit realizations exist with different topological structures and device parameters. This one-to-many mapping phenomenon reflects the inherent design flexibility in analog circuit synthesis and motivates our generative modeling approach to capture the diverse solution space.

\section{Details about datasets}
\label{appd:sec:b}
\par
\begin{figure}[ht!]
\centering
\centerline{\includegraphics[width=0.6\columnwidth]{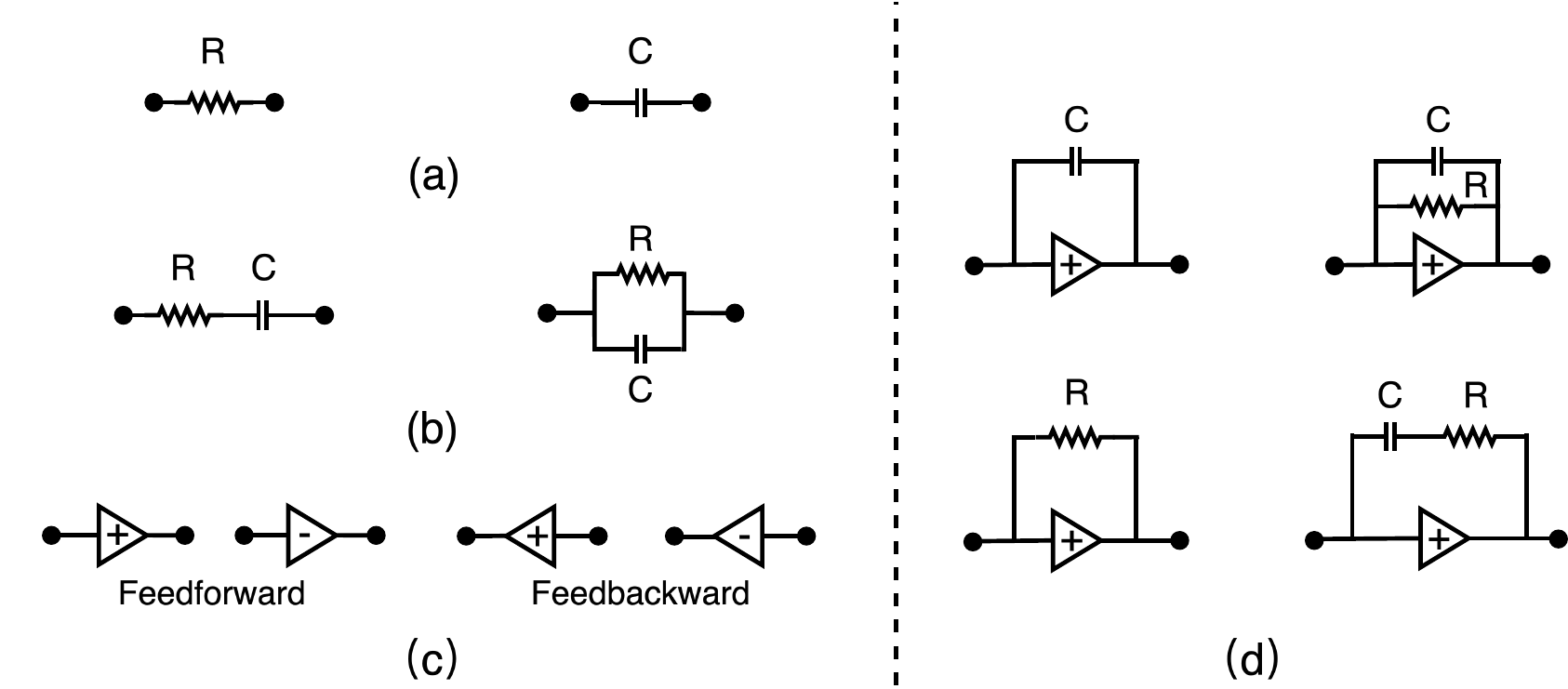}}
\caption{Hierarchical subgraph construction methodology for circuit topology representation. The decomposition strategy identifies functional blocks at multiple levels of abstraction, from elementary components to composite amplifier structures. Adapted from CktGNN~\cite{cktgnn}.}
\label{fig:subg_cons}
\end{figure}

\begin{figure}[ht!]
\centering
\centerline{\includegraphics[width=0.9\columnwidth]{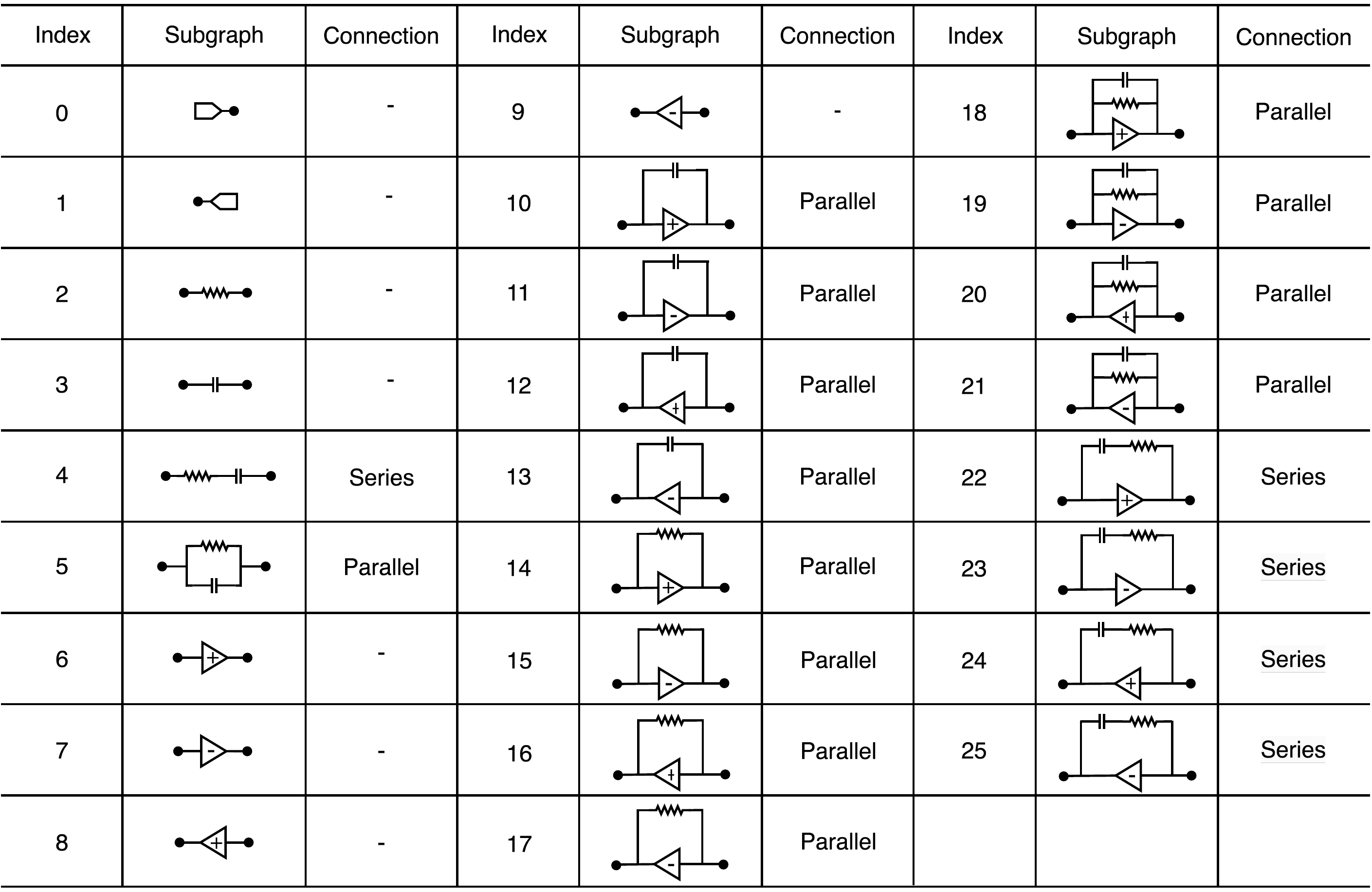}}
\caption{Complete taxonomy of the 26 subgraph types used in circuit topology representation. Each subgraph type represents a distinct functional block with specific structural and electrical characteristics, enabling compositional circuit generation.}
\label{fig:subg_table}
\end{figure}

 The circuit topology representation adopts a hierarchical subgraph-based encoding scheme, where each node in the directed acyclic graph~(DAG) corresponds to a functional block. As depicted in Fig.~\ref{fig:subg_cons}, these blocks are categorized into four hierarchical types based on their structural complexity: (a)~elementary passive components, comprising individual resistors or capacitors~(2~types); (b)~basic passive networks, consisting of resistors or capacitors connected in parallel or series configurations~(2~types); (c)~single-stage amplifiers with varying polarities~(positive, negative) and feedback configurations~(forward, feedback), yielding 4 distinct types; and (d)~composite amplifier structures that integrate single-stage amplifiers with parallel or series passive components~(16~types). This hierarchical taxonomy yields a total of 26 subgraph types, as comprehensively illustrated in Fig.~\ref{fig:subg_table}. Such a representation enables our model to capture both low-level component interactions and high-level functional blocks within the circuit topology.

\newpage

\section{More details on the evaluation metrics}
\label{appd:sec:c}
In our experimental evaluation, we employ the following metrics to assess model performance:

\begin{itemize}
  \item Retrieval precision: For a batch of circuit and specification latent vectors $\bm{z}^\text{ckt}$, $\bm{z}^\text{spec}$, we first calculate the cosine similarity matrix:
  \begin{small}
    \begin{equation}
    \label{eq_retr}
    \bm{R} (\bm{z}^\text{ckt}, \bm{z}^\text{spec}) 
    = \frac{\bm{z}^\text{ckt} \cdot \bm{z}^\text{spec}}{||\bm{z}^\text{ckt}|| \cdot || \bm{z}^\text{spec} ||}
    \end{equation}
  \end{small}%
\noindent where $\bm{R}$ denotes the cosine similarity matrix, and $||\cdot||$ is the Euclidean norm. 
  We then sort each row of the similarity matrix and take the indices $\sI_k$ of the top-k elements in order of similarity from greatest to least. Subsequently, the top-k retrieval precision is calculated as follows:
  \begin{small}
  \begin{equation}
    \label{eq_retr_acc}
    \text{Top-k} = \frac{\sum_{i=1}^{N}\left|\sI_k \cap \sG_k \right|/ k}{M}
  \end{equation}
  \end{small}%
\noindent where $M$ denotes the batch size, $\sG_k$ is the set of ground-truth indices, $|\cdot|$ denotes the cardinality of the set~(\ie, the number of elements in the set), and $\sI_k \cap \sG_k$ is the intersection of the top-k search results with relevant specifications to circuits.

  \item Specification accuracy~(Spec-Acc): This metric quantifies the proportion of generated circuits where the evaluator's estimated specifications match the original conditions. Specifically, for a batch of generated circuits with size $N$, we use our pre-trained surrogate model to encode them into latent vectors $\bm{z}^{\text{ckt}}$. Surrogate model predicts three specification metrics based on these latent vectors, denoted as $\bm{s}_{\text{Gain}}^{\prime}$, $\bm{s}_{\text{BW}}^{\prime}$, $\bm{s}_{\text{PM}}^{\prime}$. The accuracy is then computed as the ratio of correct predictions to the total number of data points:
  \begin{small}
  \begin{equation}
      \text{Spec-Acc}
      =
      \frac{
        \sum 1\left(\bm{s}_{\text{Gain}}^{\prime}, \bm{s}_{\text{Gain}}\right) 
        \wedge 1\left(\bm{s}_{\text{BW}}^{\prime}, \bm{s}_{\text{BW}}\right) 
        \wedge 1\left(\bm{s}_{\text{PM}}^{\prime}, \bm{s}_{\text{PM}}\right)}{N}
  \end{equation}
  \end{small}%
  \noindent where $1(\cdot, \cdot)$ is a function that returns 1 if the predicted class matches the ground truth and 0 otherwise.

  \item Multimodal distance~(MM-D)~\cite{t2m_gpt}: This metric evaluates the consistency between the generated circuit and the specification by calculating the average cosine distance between their encoded latent vectors. For a batch of latent vectors of encoded circuits $\bm{z}^\text{ckt}$ and their corresponding specifications $\bm{z}^\text{spec}$, we compute the cosine similarity to obtain $\bm{R}(\bm{z}^\text{ckt}, \bm{z}^\text{spec})$. The cosine distance is then obtained by subtracting $\bm{R}(\bm{z}^\text{ckt}, \bm{z}^\text{spec})$ from 1.

  \item Frechet inception distance~(FID)~\cite{fid}: To compute this metric, we first generate circuits for each joint specification class, then randomly select a circuit from the ground truth. Subsequently, we stack all the latent vectors of the generated circuits and ground truth and compute the metric value. FID evaluates the distributional differences between the generated circuits and the ground truth in the latent space.
  
  \item Diversity: To measure diversity, in each generation round, we randomly select 100 pairs of generated circuits with different joint specification classes and compute the average Euclidean distance between them to obtain the inter-class diversity~(Inter-D). Additionally, after all generations finish, we randomly select 100 pairs of circuits with the same joint specification class and obtain the intra-class diversity~(Intra-D).
  
  \item Validity: This metric measures the proportion of generated circuits with a single input and output node, free of cycles, and without feedback paths in the main path~\cite{cktgnn}.
  
  \item Novelty: This metric quantifies the proportion of decoded DAGs that have not appeared in the training dataset. For fair comparison, we scale the mean and variance parameters of the distribution to align with those of the training dataset, as done in CktGNN~\cite{cktgnn}.
  \item 
    Reconstruction accuracy~(Recon-Acc): We first encode the ground truth circuits into latent representations using the reparameterization trick~\cite{vae} and decode them back into DAGs. This metric is calculated as the proportion of the decoded DAGs that are identical to the ground truth.
  
\end{itemize}

\section{Additional experiments on training epochs}
\label{appd:sec:d}
\par We conduct an ablation study on training epochs to investigate the impact of training duration on model performance for CktGen, CVAEGAN, and LDT. Results are summarized in Tables~\ref{tab:epoch_cktgen}, \ref{tab:epoch_cvaegan}, and \ref{tab:epoch_ldt}.

\par As demonstrated in Table~\ref{tab:epoch_cktgen}, CktGen exhibits significant performance improvements as training progresses from 100 to 700 epochs. The model reaches its optimal performance at 600 epochs, achieving substantial gains in retrieval precision (exceeding 35\%) and specification accuracy (approaching 48\%), while maintaining high diversity and validity. These results indicate that extended training enables CktGen to better align generated circuits with target specifications while preserving generation diversity.

\par In contrast, both baseline models exhibit limited improvements with additional training. CVAEGAN~(Table~\ref{tab:epoch_cvaegan}) demonstrates consistently low retrieval precision across all training epochs, with specification accuracy remaining below 0.7\%. Similarly, LDT~(Table~\ref{tab:epoch_ldt}) shows stable but modest performance, with retrieval precision and specification accuracy both below 2.5\%. Although both baseline models achieve high validity rates exceeding 97\%, their overall performance remains substantially inferior to CktGen, suggesting that baseline architectures struggle to effectively model the specification-conditioned generation task even with prolonged training. These findings confirm the effectiveness of CktGen's architecture and training strategies in learning robust, specification-aligned circuit generation.

\begin{table*}[t!]
    \centering
    \begin{scriptsize}
        \caption{Ablation study on training epochs for CktGen on Ckt-Bench-101.}
        \begin{tabular}{lccccccccc}
        \hline
        & \multicolumn{3}{c}{Retrieval Precision (\%)} & & & & & \\ 
        \cline{2-4}
        
        \multirow{-1.9}{*}{Epoch} 
        & Top-1
        & Top-2
        & Top-3
        & \multirow{-1.9}{*}{Spec-Acc (\%)} 
        & \multirow{-1.9}{*}{MM-D}
        & \multirow{-1.9}{*}{FID} 
        & \multirow{-1.9}{*}{Inter-D}
        & \multirow{-1.9}{*}{Intra-D}
        & \multirow{-1.9}{*}{Validity (\%)}

        \\ \hline
        
        100
        & 4.776
        & 8.823
        & 12.66
        & 4.223
        & 0.580
        & 8.760
        & 7.699
        & 2.280
        & 85.33
        \\
    
        200
        & 11.81
        & 19.74
        & 25.40
        & 13.30
        & 0.501
        & 6.415
        & 8.232
        & 2.410
        & 90.85
        \\
    
        300
        & 24.32
        & 39.83
        & 48.31
        & 33.24
        & 0.430
        & \textbf{5.318}
        & 8.269
        & 2.211
        & 89.76
        \\
        
        400
        & 29.69
        & 46.27
        & 55.58
        & 39.65
        & \underline{0.401}
        & \underline{5.333}
        & 8.554
        & 1.810
        & 91.48
        \\
    
        500
        & \underline{34.60}
        & \underline{50.94}
        & 58.99
        & \underline{40.66}
        & 0.404
        & 5.496
        & 8.518
        & 1.716
        & \textbf{95.98}
        \\    
    
        \rowcolor{aliceblue}
        600
        & \textbf{35.73}
        & \textbf{55.93}
        & \textbf{65.21}
        & \textbf{47.57}
        & \textbf{0.385}
        & 6.092
        & \textbf{8.574}
        & 1.987
        & \underline{95.47}
        \\

        700
        & 33.46
        & 50.38
        & \underline{59.10}
        & 39.14
        & 0.402
        & 6.277
        & \underline{8.591}
        & 2.060
        & 93.38
        \\
        
        \hline
        \end{tabular}
        \label{tab:epoch_cktgen}
        \end{scriptsize}
\end{table*}

\begin{table*}[t!]
\centering
\begin{scriptsize}
    \caption{Ablation study on training epochs for CVAEGAN on Ckt-Bench-101.}
    \begin{tabular}{lccccccccc}
    \hline
    & \multicolumn{3}{c}{Retrieval Precision (\%)} & & & & & \\ 
    \cline{2-4}
    
    \multirow{-1.9}{*}{Epoch} 
    & Top-1
    & Top-2
    & Top-3
    & \multirow{-1.9}{*}{Spec-Acc (\%)} 
    & \multirow{-1.9}{*}{MM-D} 
    & \multirow{-1.9}{*}{FID} 
    & \multirow{-1.9}{*}{Inter-D}
    & \multirow{-1.9}{*}{Intra-D}
    & \multirow{-1.9}{*}{Validity (\%)} 
    
    \\ \hline
    
    \rowcolor{aliceblue}
    100
    & \textbf{0.789}
    & \textbf{1.468}
    & \textbf{2.179}
    & 0.635
    & \textbf{0.865}
    & \underline{4.782}
    & \underline{7.575}
    & 7.362
    & \underline{98.05}
    \\

    200
    & 0.666
    & 1.264
    & 1.899
    & 0.616
    & \underline{0.886}
    & 4.841
    & \textbf{7.599}
    & 7.436
    & \textbf{98.13}
    \\

    300
    & \underline{0.729}
    & \underline{1.336}
    & \underline{1.908}
    & \underline{0.669}
    & 0.888
    & \textbf{4.771}
    & 7.549
    & 7.407
    & 97.84
    \\
    
    400
    & 0.720
    & 1.311
    & 1.896
    & \underline{0.669}
    & 0.888
    & 4.833
    & 7.544
    & 7.405
    & 98.00
    \\

    500
    & 0.713
    & 1.311
    & 1.883
    & \underline{0.669}
    & 0.888
    & 4.832
    & 7.543
    & 7.405
    & 98.01
    \\    

    600
    & 0.717
    & 1.323
    & 1.899
    & \textbf{0.673}
    & 0.888
    & 4.828
    & 7.545
    & 7.406
    & 98.00
    \\
    
    700
    & 0.713
    & 1.327
    & \underline{1.908}
    & 0.666
    & 0.888
    & 4.832
    & 7.543
    & 7.406
    & 98.03
    \\
    
    \hline
    \end{tabular}
    \label{tab:epoch_cvaegan}
    \end{scriptsize}
\end{table*}

\begin{table*}[t!]
    \centering
    \begin{scriptsize}
        \caption{Ablation study on training epochs for LDT on Ckt-Bench-101.}
        \begin{tabular}{lccccccccc}
        \hline
        & \multicolumn{3}{c}{Retrieval Precision (\%)} & & & & & \\ 
        \cline{2-4}
        
        \multirow{-1.9}{*}{Epoch} 
        & Top-1
        & Top-2
        & Top-3
        & \multirow{-1.9}{*}{Spec-Acc (\%)} 
        & \multirow{-1.9}{*}{MM-D} 
        & \multirow{-1.9}{*}{FID} 
        & \multirow{-1.9}{*}{Inter-D}
        & \multirow{-1.9}{*}{Intra-D}
        & \multirow{-1.9}{*}{Validity (\%)} 
        
        \\ \hline
        
        \rowcolor{aliceblue}
        100
        & 2.371
        & \textbf{4.981}
        & 7.471
        & 1.915
        & \underline{0.598}
        & 30.72
        & \textbf{11.26}
        & 7.200
        & \textbf{98.32}
        \\
    
        200
        & \textbf{2.493}
        & 4.949
        & \underline{7.515}
        & \underline{2.047}
        & \textbf{0.596}
        & \textbf{30.24}
        & \underline{11.16}
        & 7.189
        & \underline{98.30}
        \\
    
        300
        & \underline{2.490}
        & 4.962
        & 7.506
        & 2.040
        & \textbf{0.596}
        & \underline{30.30}
        & \underline{11.16}
        & 7.190
        & 98.29
        \\
        
        400
        & 2.481
        & \underline{4.971}
        & \textbf{7.518}
        & 2.047
        & \textbf{0.596}
        & 30.36
        & \underline{11.16}
        & 7.189
        & 98.27
        \\
    
        500
        & 2.484
        & 4.965
        & 7.509
        & \textbf{2.050}
        & \textbf{0.596}
        & 30.38
        & \underline{11.16}
        & 7.189
        & 98.28
        \\    
    
        600
        & 2.474
        & 4.956
        & 7.493
        & 2.044
        & \textbf{0.596}
        & 30.40
        & \underline{11.16}
        & 7.190
        & 98.27
        \\
        
        \hline
        \end{tabular}
        
        \label{tab:epoch_ldt}
        \end{scriptsize}
\end{table*}


\end{document}